\documentclass{article}

\usepackage{microtype}
\usepackage{graphicx}
\usepackage{subfigure}
\usepackage{booktabs} 

\usepackage{hyperref}

\usepackage[accepted]{icml2024}

\usepackage[textsize=tiny]{todonotes}

\usepackage{microtype}
\usepackage{subfigure}
\usepackage{booktabs} 

\usepackage{hyperref}

\usepackage{amsmath}
\usepackage{amssymb}
\usepackage{mathtools}
\usepackage{amsthm}
\usepackage{xspace}

\usepackage[capitalize,noabbrev]{cleveref}

\theoremstyle{plain}

\theoremstyle{definition}

\theoremstyle{remark}

\usepackage[textsize=tiny]{todonotes}

\usepackage{hyperref}
\usepackage{url}
\usepackage{tikz}
\usepackage{multirow}
\usepackage{xspace}
\usepackage{booktabs}
\usepackage{ulem}

\usepackage{graphicx} 

\usepackage{caption}
\usepackage{adjustbox}

\usepackage[flushleft]{threeparttable}
\usepackage{enumitem}
\usepackage{verbatim}
\usepackage{listings}
\usepackage{xcolor}
\usepackage{amssymb}
\usepackage{scalerel} 
\usepackage{float}
\usepackage{subcaption}
\usepackage[inkscapeformat=png]{svg}
\usepackage{bbding}
\usepackage{comment}
\usepackage{tabularray}
\usepackage{booktabs,tabularx}
\usepackage{stfloats}

\newcommand{\approach}{\texttt{CodeAct}\xspace}
\newcommand{\dataname}{CodeActInstruct\xspace}
\newcommand{\modelname}{CodeActAgent\xspace}
\newcommand{\evalname}{M$^3$ToolEval\xspace}

\newcommand{\sref}[1]{\S\ref{#1}}
\newcommand{\fref}[1]{Fig.~\ref{#1}}
\newcommand{\tref}[1]{Tab.~\ref{#1}}
\renewcommand{\eqref}[1]{Eq.~\ref{#1}}

\newcommand{\redwrong}{{\textcolor{red}{\XSolidBrush}}}
\definecolor{ao(english)}{rgb}{0.0, 0.5, 0.0}
\newcommand{\greencheck}{{\textcolor{ao(english)}{\CheckmarkBold}}}

\newcommand{\interalia}[1]{\citep[\textit{inter alia}]{#1}}

\NewDocumentCommand{\hao}
{ mO{} }{\textcolor{blue}{\textsuperscript{\textit{Hao}}\textsf{\textbf{\small[#1]}}}}

\NewDocumentCommand{\heng}
{ mO{} }{\textcolor{red}{\textsuperscript{\textit{Heng}}\textsf{\textbf{\small[#1]}}}}

\NewDocumentCommand{\xingyao}
{ mO{} }{\textcolor{orange}{\textsuperscript{\textit{Xingyao}}\textsf{\textbf{\small[#1]}}}}

\NewDocumentCommand{\yy}
{ mO{} }{\textcolor{pink}{\textsuperscript{\textit{coolYY}}\textsf{\textbf{\small[#1]}}}}

\NewDocumentCommand{\lifan}
{ mO{} }{\textcolor{cyan}{\textsuperscript{\textit{Lifan}}\textsf{\textbf{\small[#1]}}}}

\NewDocumentCommand{\yizhe}
{ mO{} }{\textcolor{green}{\textsuperscript{\textit{Yizhe}}\textsf{\textbf{\small[#1]}}}}

\NewDocumentCommand{\yunzhu}
{ mO{} }{\textcolor{magenta}{\textsuperscript{\textit{Yunzhu}}\textsf{\textbf{\small[#1]}}}}

\usepackage{microtype}
\usepackage{hyperref}
\hypersetup{
	colorlinks=true,
	linkcolor=cyan,
	filecolor=blue,      
	urlcolor=cyan,
	citecolor=cyan,
}

\icmltitlerunning{Executable Code Actions Elicit Better LLM Agents}

\begin{document}

\twocolumn[
\icmltitle{
Executable Code Actions Elicit Better LLM Agents
}




\begin{icmlauthorlist}
\icmlauthor{Xingyao Wang}{uiuc}
\icmlauthor{Yangyi Chen}{uiuc}
\icmlauthor{Lifan Yuan}{uiuc}
\icmlauthor{Yizhe Zhang}{apple}
\icmlauthor{Yunzhu Li}{uiuc}
\icmlauthor{Hao Peng}{uiuc}
\icmlauthor{Heng Ji}{uiuc}
\end{icmlauthorlist}

\icmlaffiliation{uiuc}{Department of Computer Science, University of Illinois Urbana-Champaign}
\icmlaffiliation{apple}{Apple}
\icmlcorrespondingauthor{Xingyao Wang}{xingyao6@illinois.edu}
\icmlcorrespondingauthor{Heng Ji}{hengji@illinois.edu}

\icmlkeywords{Machine Learning, ICML}

\vskip 0.3in
]



\printAffiliationsAndNotice{}  

\begin{abstract}
\looseness=-1
Large Language Model (LLM) agents, capable of performing a broad range of actions, such as invoking tools and controlling robots, show great potential in tackling real-world challenges.
LLM agents are typically prompted to produce actions by generating JSON or text in a pre-defined format, which is usually limited by constrained action space (e.g., the scope of pre-defined tools) and restricted flexibility (e.g., inability to compose multiple tools).
This work proposes to use executable Python \textbf{code} to consolidate LLM agents' \textbf{act}ions into a unified action space (\approach).
Integrated with a Python interpreter, \approach can execute code actions and dynamically revise prior actions or emit new actions upon new observations through multi-turn interactions.
Our extensive analysis of 17 LLMs on API-Bank and a newly curated benchmark shows that \approach outperforms widely used alternatives (up to 20\% higher success rate).
The encouraging performance of \approach motivates us to build an open-source LLM agent that interacts with environments by executing interpretable code and collaborates with users using natural language.
To this end, we collect an instruction-tuning dataset \dataname that consists of 7k multi-turn interactions using \approach.
We show that it can be used with existing data to improve models in agent-oriented tasks without compromising their general capability.
\modelname, finetuned from Llama2 and Mistral, is integrated with Python interpreter and uniquely tailored to perform sophisticated tasks (e.g., model training) using existing libraries and autonomously self-debug\footnote{The code, data, model, and demo are available at \url{https://github.com/xingyaoww/code-act}.}.
\vspace{-0.2cm}
\end{abstract}
\section{Introduction}
\looseness=-1
\begin{figure*}[!htb]
    \centering
    \includegraphics[width=\textwidth]{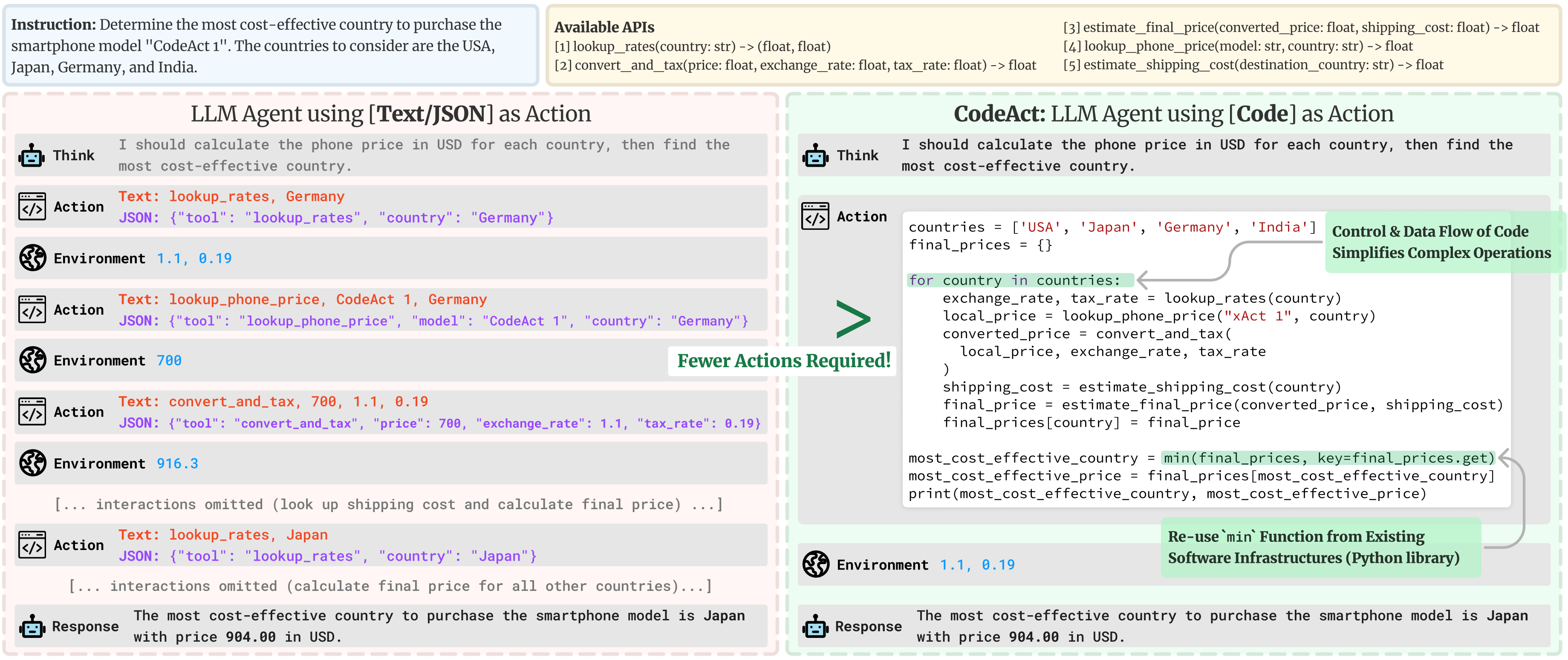}
    \vspace{-5pt}
    \includegraphics[width=\textwidth]{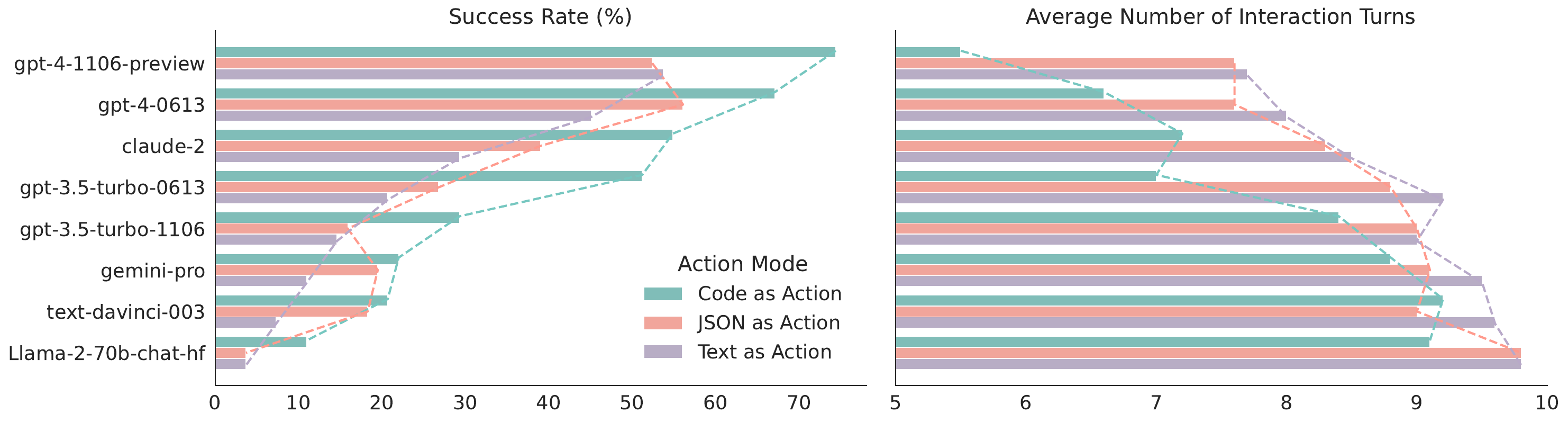}
    \vspace{-15pt}
    \caption{Comparison between \approach and Text / JSON as action. \textbf{(top)} Illustrative example comparing different actions. \textbf{(bottom)} Quantitative results on \evalname (\sref{sec:control-data-flow}).}
    \label{fig:illustrative_example}
    \label{fig:zeroshot_acteval} 
\end{figure*}

Large Language Models (LLMs) have emerged as a pivotal breakthrough in natural language processing (NLP).
When augmented with \textit{action} modules that allow access to APIs, their action space expands beyond conventional text processing, allowing LLMs to acquire capabilities such as tool invocation and memory management~\citep{mialon2023augmented, schick2023toolformer} and venture into real-world tasks such as controlling robots~\citep{saycan2022arxiv,huang2023voxposer,ma2023eureka} and performing scientific experiments~\citep{bran2023chemcrow}.

We inquire: \textit{how to effectively expand LLM agents' action space for solving complex real-world problems?} 
Much existing research has examined using text~\interalia{yao2022react,park2023generative} or JSON~\interalia{Qin2023ToolLLMFL,Chase_LangChain_2022} to produce actions (e.g., tool uses in \fref{fig:illustrative_example} top left).
However, both methods typically suffer from constrained scope of action spaces (actions are usually tailored for specific tasks) and restricted flexibility (e.g., inability to compose multiple tools in a single action).
As an alternative approach, several work \citep{codeaspolicies2022,progprompt,wang2023voyager} demonstrate the potential of using LLMs to generate code to control robots or game characters.
However, they typically rely on pre-specified control primitives and hand-engineered prompts and, more importantly, struggle to dynamically adjust or emit actions based on new environmental observation and feedback.

This work proposes \approach, a general-purpose framework that allows LLMs to generate executable Python \textbf{code} as \textbf{act}ions (\fref{fig:illustrative_example} top right).
\approach is designed to handle a variety of applications and comes with unique advantages:
%
\begin{itemize}[noitemsep,topsep=0pt,parsep=0pt,partopsep=0pt,leftmargin=18pt]
\item[(1) ]Integrated with a Python interpreter, \approach can execute code actions and \textit{dynamically adjust prior actions or emit new action} based on observations (e.g., code execution results) it receives through multiple turns of interactions.

\item[(2)]
Code actions allow LLM to leverage existing \textit{software packages}. \approach can use readily available Python packages for an expanded action space instead of hand-crafted task-specific tools \citep{Yuan2023CRAFTCL,shen2023hugginggpt}. It also allows LLM to use automated feedback (e.g., error messages) implemented in most software to improve task-solving by self-debugging its generated code \citep{chen2023teaching,Wang2023LeTI}.

\item[(3)] 
\textit{Code data} is widely used in pre-training today's  LLMs \citep{yang2024llm}.
These models are already familiar with structured programming languages, allowing cost-effective adoption of \approach.

\item[(4)]
Compared to JSON and text with a pre-defined format, code inherently supports \textit{control and data flow}, allowing for the storage of intermediate results as variables for reuse and the composition of multiple tools to perform complex logical operations (e.g., if-statements, for-loops) with \textit{one} piece of code, thereby unlocking LLMs' potential to tackle complex tasks by leveraging its pre-trained knowledge of programming.
In \fref{fig:illustrative_example}, an LLM using with \approach (top right) can apply the same sequence of tools (e.g., passing one tool's output as input to another tool using the data flow feature) to \textit{all} inputs through for-loops (i.e., control flow feature) with \textit{one} action; while text or JSON have to take action for every input (top left).
\end{itemize}

Our extensive experiments with 17 LLMs (including both open-source and proprietary ones) confirm the above benefits (3 
\& 4) of \approach.
To demonstrate benefit (3), our first experiment (\sref{sec:data-factor-format}) compares \approach to baselines on basic tasks involving \textit{atomic tool use} (i.e., only one tool is used per action), ablating the control and data flow advantage offered by \approach.
The results show that, for most LLMs, \approach achieves comparable or better performance than the baselines.
\approach's performance gains are more prominent on complex tasks, as demonstrated in our second experiment (benefit 4).
We curate a new benchmark consisting of 82 human-curated tasks that typically require \textbf{m}ultiple calls to \textbf{m}ultiple tools in \textbf{m}ulti-turn interactions (\evalname; \sref{sec:control-data-flow}).
Problems in this benchmark often require intricate coordination and composition of multiple tools.
With its strengths in control and data flow, \approach achieves up to a 20\% absolute improvement over baselines on the success rate of solving the problems while requiring up to 30\% fewer actions.
These performance gains widen as the capabilities of the LLMs increase (\fref{fig:illustrative_example} bottom).

\begin{figure*}[!th]
    \centering
    \includegraphics[width=\textwidth]{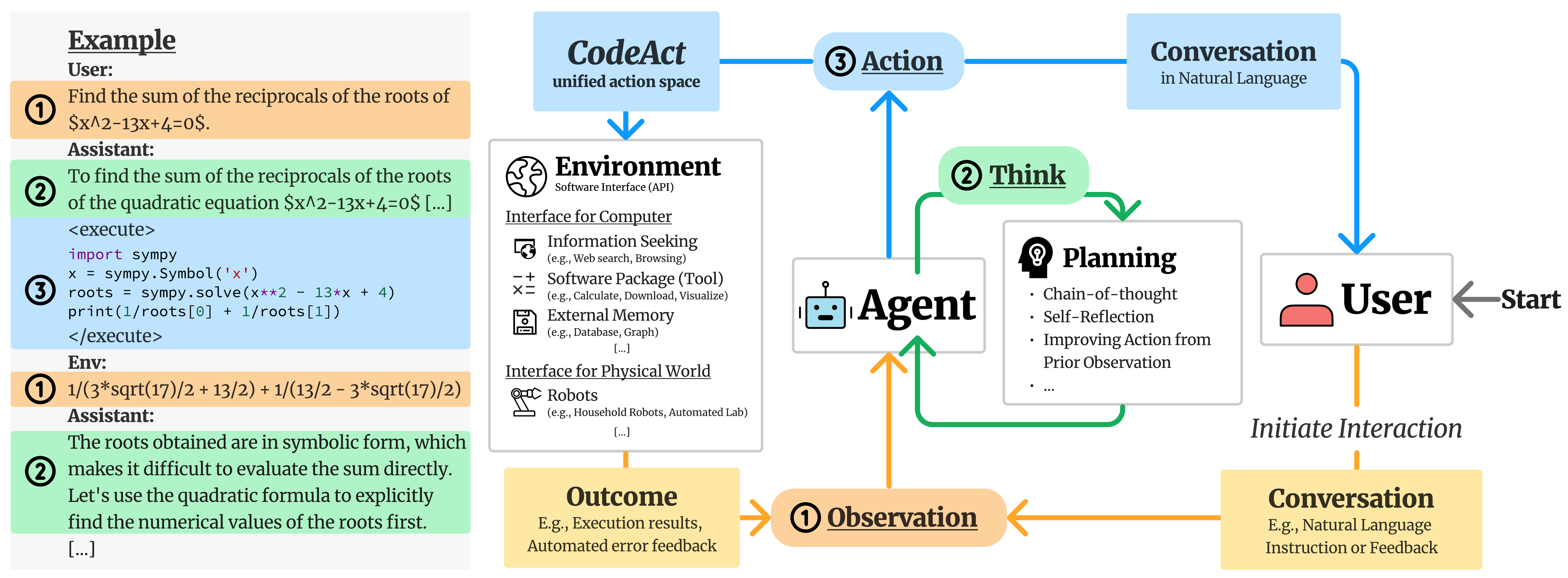}
    \vspace{-20pt}
    \caption{General agent multi-turn interaction framework that describes the role of \approach and motivates the construction of our data mixture. \dataname focuses on the \textit{agent-environment} interactions and specifically filters for the self-improved \textit{planning} behavior, while general conversation data we include focuses on \textit{agent-user} interaction (\sref{sec:agent_env_data}).
    }
    \label{fig:llm_agent_framework}
\end{figure*}

The promising performance of \approach motivates an open-source LLM agent that can effectively act through \approach, and collaborate with humans through natural language.
To this end, we collect an instruction-tuning dataset \dataname consisting of 7k high-quality multi-turn interaction trajectories with \approach (\sref{sec:agent_env_data}).
\dataname is motivated by a general agent framework consisting of agent, user, and environments (\fref{fig:llm_agent_framework}) and focuses on agent-environment interactions with the computer (information seeking, software package use, external memory) and the physical world (robot planning).
On \dataname, we perform careful data selection to promote the capability of improving from multi-turn interaction (e.g., self-debug).
We show that \dataname can be used with commonly used instruction tuning data to improve the models' performance in agent tasks without compromising their general capabilities (e.g., knowledge-based QA, coding, instruction following, \sref{sec:llm_agent_evaluation}).
Our model, dubbed \modelname, is finetuned from LLaMA-2 \citep{touvron2023llama} and Mistral-7B \citep{jiang2023mistral} and improves on out-of-domain agent tasks with not only \approach, but also text action in a pre-defined format (\sref{sec:llm_agent_evaluation}).

\approach can further benefit from multi-turn interactions and existing software (benefit 1 \& 2, \sref{sec:multiturn_software_benefit}).
As shown in \fref{fig:qualitative_example}, \modelname, designed for seamless integration with Python, can carry out sophisticated tasks (e.g., model training, data visualization) using existing Python packages. Error messages from the environment further enable it to rectify errors autonomously through self-debugging in multi-turn interaction.
Thanks to LLM's extensive programming knowledge acquired during pre-training, these are achieved without needing in-context demonstrations, reducing the human efforts for adapting \modelname to different tasks.

\begin{table*}[!t]
\centering
\caption{
The benefit of \approach compared to using Text/JSON for LLM action.
}
\vspace{-10pt}
\resizebox{\textwidth}{!}{
\begin{threeparttable}
\begin{tabular}{@{} p{5.3cm} p{7.5cm} p{7cm} @{}}
\toprule
 & \textbf{\approach for LLM action} & \textbf{JSON or Text for LLM action} \\
\midrule
\textbf{Availability of Data} & \greencheck Large quantity of code available$^1$ for pre-training & \redwrong Data curation required for particular format \\
\midrule
\textbf{Complex Operation} (e.g., looping, composition of multiple tools) &  \multirow{2}*{\greencheck Natively supported via control and data flow} & \redwrong Requires careful engineering if feasible (e.g., define new tools to mimic if-statement) \\
\midrule
\multirow{2}*{\textbf{Availability of Tools}} & \multirow{2}*{\greencheck Can directly use existing software packages$^2$} & \redwrong Requires human effort to curate tools from scratch or existing software \\
\midrule
\multirow{3}*{\textbf{Automated Feedback}} & \greencheck Feedback mechanism$^3$ (e.g., traceback) is already implemented as an infrastructure for most programming languages & \redwrong Requires human effort to provide feedback or re-route feedback from the underlying programming language used to implement the tools \\

\bottomrule
\end{tabular}
\begin{tablenotes}
    \item[1] Including code demonstrating useful behaviors for LLM agents (e.g., task decomposition, coordination of multiple function calls to different tools).
    \item[2] Human-written Python packages covering a wide range of applications are available on \url{https://pypi.org/}.
    \item[3] For example, in Python, errors and exceptions (\url{https://docs.python.org/3/tutorial/errors.html}) are available. Most software provides error messages in natural language to help human programmers debug their code. \approach enables LLM to use them directly.
\end{tablenotes}
\end{threeparttable}
}
\label{tab:codeact_comparison_table}
\vspace{-14pt}
\end{table*}
\everypar{\looseness=-1}
\section{\approach Makes LLMs Better Agents}

In this section, we first describe \approach framework (\sref{sec:codeact_definition}) and provide empirical evidence that supports the choice of \approach.
We focus on Python as the programming language for \approach due to its popularity (ranked top-1 at \cite{tiobe}) and numerous open-source packages. 
We aim to answer several research questions (RQs) using 17 off-the-shelf LLMs. In \sref{sec:data-factor-format}, we examine RQ1: Does LLMs' familiarity with code due to a large amount of code pre-training data bring \approach advantages over text and JSON?
We discuss RQ2 in \sref{sec:control-data-flow}: Does \approach benefit from Python’s innate control and data flow feature in complex problems?
Finally, as an additional benefit, we discuss how using \approach further enhances LLM agents by enabling multi-turn interactions and allowing them to access existing software in \sref{sec:multiturn_software_benefit} and \fref{fig:qualitative_example}.

\subsection{What is \approach?}
\label{sec:codeact_definition}

In \fref{fig:llm_agent_framework}, we first introduce a general multi-turn interaction framework for LLM agents' real-world usage that considers three roles \citep{yang2024unified}: \textit{agent}, \textit{user}, and \textit{environment}.
We define \textit{interaction} as the information exchange between the agent and an external entity (user or environment).
For each turn of interaction, the agent receives an \textit{observation} (input) either from the user (e.g., natural language instruction) or the environment (e.g., code execution result), optionally planning for its action through chain-of-thought \citep{wei2022chain}, and emits an \textit{action} (output) to either user in natural language or the environment.
\approach employs Python code to consolidate all \textit{actions} for \textit{agent-environment interaction}. In \approach, each emitted \textit{action} to the environment is a piece of Python code, and the agent will receive outputs of code execution (e.g., results, errors) as \textit{observation}.
We include an example prompt of \approach in \sref{sec:codeact_prompt}.

\subsection{\approach Shows the Promise as a Strong Tool Use Framework}
\label{sec:data-factor-format}
In this section, we perform a controlled experiment to understand which format (text, JSON, \approach) is more likely to lead an LLM to generate correct \textit{atomic} tool calls.
The performance in this experiment reflects LLM's familiarity with the corresponding format.
We hypothesize that using \approach to call tools is a more natural way to use tools for the models, which typically have extensive exposure to \textit{code data} during their training.

\noindent \textbf{Setup.}
We re-purpose API-Bank \citep{li2023apibank} and test LLMs' API-calling performance, comparing \approach, JSON, and text actions.
For each evaluation instance, we instruct LLM to generate \textit{one atomic} tool call in the format of a Python function call, JSON object, or text expression in a pre-defined format. A concrete example is shown in \tref{tab:apibank-action-example}.
We use API-Bank's level-1 instructions and the provided toolset. To evaluate API-calling, we follow their \textit{correctness} metric, matching the ground-truth API outputs with the actual model-generated API's execution outputs.

\noindent \textbf{Results.}
We present results in \tref{tab:api-bank-results}. 
For most LLMs, \approach achieves comparable or better performance even in atomic actions (the simplistic tool use scenario) where its control and data flow strengths are ablated.
Compared to closed-source LLMs, \approach's improvements are more prominent in open-source models. Furthermore, code data is usually more accessible for fine-tuning open-source LLMs than the specialized JSON or text tool-calling format.
Although JSON is consistently weaker than other approaches for open-source models, it achieves decent performance with closed-source LLMs, indicating that these closed-source models may have gone through targeted fine-tuning toward their JSON capabilities.
These results suggest optimizing for \approach is a better route for open-source LLMs than alternatives to improve their tool-use capabilities, as they already show good initial \approach capability due to extensive exposure to code data during pre-training.

\begin{table*}[bpth]
\begin{minipage}{0.40\textwidth}
\centering
\caption{
Atomic API call correctness on API-Bank.
The best performance is \textbf{bolded}, and the second-best is \underline{underlined}.
}
\vspace{-5pt}
\label{tab:api-bank-results}
\begin{adjustbox}{max width=\textwidth}
\begin{tabular}{@{} l rrr @{}}
\toprule
& \multicolumn{3}{c}{\textbf{Correctness} (\%, $\uparrow$)} \\
\textbf{Format of Action} &  \approach &  JSON &  Text \\
\midrule
\multicolumn{4}{c}{\textit{Open-source LLMs}} \\

\texttt{CodeLlama-7b-Instruct-hf}  &  \underline{$12.5$} &              $12.0$ &     $\mathbf{17.0}$ \\
\texttt{CodeLlama-13b-Instruct-hf} &  \underline{$11.8$} &               $7.8$ &     $\mathbf{14.0}$ \\
\texttt{CodeLlama-34b-Instruct-hf} &     $\mathbf{17.3}$ &              $12.0$ &  \underline{$16.8$} \\
\texttt{Llama-2-7b-chat-hf}        &     $\mathbf{28.8}$ &              $11.3$ &  \underline{$25.8$} \\
\texttt{Llama-2-13b-chat-hf}       &     $\mathbf{38.1}$ &               $8.5$ &  \underline{$37.3$} \\
\texttt{Llama-2-70b-chat-hf}       &  \underline{$35.6$} &              $14.3$ &     $\mathbf{37.6}$ \\
\texttt{Mistral-7B-Instruct-v0.1}  &   \underline{$2.5$} &               $2.3$ &      $\mathbf{3.0}$ \\
\texttt{lemur-70b-chat-v1}         &     $\mathbf{58.6}$ &              $46.6$ &  \underline{$56.1$} \\

\midrule
\multicolumn{4}{c}{\textit{Closed-source LLMs}} \\

\texttt{claude-2}                  &     $\mathbf{76.7}$ &              $59.4$ &  \underline{$73.7$} \\
\texttt{claude-instant-1}          &     $\mathbf{75.2}$ &              $64.9$ &  \underline{$73.2$} \\
\texttt{gemini-pro}                &              $70.4$ &     $\mathbf{73.2}$ &  \underline{$71.2$} \\
\texttt{gpt-3.5-turbo-0613}        &     $\mathbf{74.4}$ &  \underline{$73.9$} &              $73.4$ \\
\texttt{gpt-3.5-turbo-1106}        &  \underline{$75.4$} &     $\mathbf{78.4}$ &              $73.4$ \\
\texttt{gpt-4-0613}                &  \underline{$75.4$} &     $\mathbf{82.0}$ &              $74.4$ \\
\texttt{gpt-4-1106-preview}        &  \underline{$76.7$} &     $\mathbf{82.7}$ &              $73.4$ \\
\texttt{text-davinci-002}          &     $\mathbf{69.2}$ &  \underline{$59.6$} &              $57.4$ \\
\texttt{text-davinci-003}          &  \underline{$75.4$} &     $\mathbf{76.9}$ &              $69.7$ \\
\midrule

\multicolumn{4}{c}{\textbf{Frequency of Best-Performing Format} $\uparrow$} \\
Open-source & $\mathbf{4}$ & $0$ & \underline{$4$} \\
Closed-source & \underline{$4$} & $\mathbf{5}$ & $0$ \\
\textbf{Overall} & $\mathbf{8}$ & $\underline{5}$ & $4$ \\
\bottomrule
\end{tabular}
\end{adjustbox}

\end{minipage}\hfill%
\begin{minipage}{0.58\textwidth}
\centering
\caption{Success rates (higher the better) and average turns required per instance (lower the better) on \evalname. The best results for each model are \textbf{bolded}, and the second-best ones are \underline{underlined}.
}
\vspace{-5pt}
\label{tab:zeroshot_act_results}
\begin{adjustbox}{max width=\textwidth}
\begin{tabular}{@{} l rrr  m{0.01em}  rrr @{}}
\toprule
{} & \multicolumn{3}{c}{\textbf{Success Rate (\%, $\uparrow$)}} && \multicolumn{3}{c}{\textbf{Avg. Turns} ($\downarrow$)} \\
\cmidrule{2-4}
\cmidrule{6-8}

\textbf{Format of Action} & \approach & JSON & Text && \approach & JSON & Text \\

\midrule

\multicolumn{8}{c}{\textit{Open-source LLMs}} \\

\texttt{CodeLlama-7b-Instruct-hf}  &      $\mathbf{4.9}$ &   \underline{$2.4$} &   \underline{$2.4$} &&     $\mathbf{9.7}$ &   \underline{$9.9$} &   \underline{$9.9$} \\
\texttt{CodeLlama-13b-Instruct-hf} &      $\mathbf{4.9}$ &      $\mathbf{4.9}$ &      $\mathbf{4.9}$ &&  \underline{$9.8$} &   \underline{$9.8$} &      $\mathbf{9.7}$ \\
\texttt{CodeLlama-34b-Instruct-hf} &      $\mathbf{2.4}$ &   \underline{$0.0$} &   \underline{$0.0$} &&     $\mathbf{9.9}$ &  \underline{$10.0$} &  \underline{$10.0$} \\
\texttt{Llama-2-7b-chat-hf}        &               $0.0$ &   \underline{$1.2$} &      $\mathbf{2.4}$ &&     $\mathbf{8.9}$ &   \underline{$9.5$} &               $9.6$ \\
\texttt{Llama-2-13b-chat-hf}       &      $\mathbf{0.0}$ &      $\mathbf{0.0}$ &      $\mathbf{0.0}$ &&     $\mathbf{9.7}$ &  \underline{$10.0$} &  \underline{$10.0$} \\
\texttt{Llama-2-70b-chat-hf}       &     $\mathbf{11.0}$ &   \underline{$3.7$} &   \underline{$3.7$} &&     $\mathbf{9.1}$ &   \underline{$9.8$} &   \underline{$9.8$} \\
\texttt{Mistral-7B-Instruct-v0.1}  &               $0.0$ &      $\mathbf{3.7}$ &   \underline{$1.2$} &&             $10.0$ &      $\mathbf{9.8}$ &   \underline{$9.9$} \\
\texttt{lemur-70b-chat-v1}         &  \underline{$13.4$} &     $\mathbf{15.9}$ &              $12.2$ &&     $\mathbf{9.1}$ &   \underline{$9.3$} &               $9.4$ \\

\midrule 
\multicolumn{8}{c}{\textit{Closed-source LLMs}} \\

\texttt{claude-2}                  &     $\mathbf{54.9}$ &  \underline{$39.0$} &              $29.3$ &&     $\mathbf{7.2}$ &   \underline{$8.3$} &               $8.5$ \\
\texttt{claude-instant-1}          &              $20.7$ &     $\mathbf{31.7}$ &  \underline{$24.4$} &&  \underline{$8.8$} &      $\mathbf{8.6}$ &               $8.9$ \\
\texttt{gemini-pro}                &     $\mathbf{22.0}$ &  \underline{$19.5$} &              $11.0$ &&     $\mathbf{8.8}$ &   \underline{$9.1$} &               $9.5$ \\
\texttt{gpt-3.5-turbo-0613}        &     $\mathbf{51.2}$ &  \underline{$26.8$} &              $20.7$ &&     $\mathbf{7.0}$ &   \underline{$8.8$} &               $9.2$ \\
\texttt{gpt-3.5-turbo-1106}        &     $\mathbf{29.3}$ &  \underline{$15.9$} &              $14.6$ &&     $\mathbf{8.4}$ &   \underline{$9.0$} &   \underline{$9.0$} \\
\texttt{gpt-4-0613}                &     $\mathbf{67.1}$ &  \underline{$56.1$} &              $45.1$ &&     $\mathbf{6.6}$ &   \underline{$7.6$} &               $8.0$ \\
\texttt{gpt-4-1106-preview}        &     $\mathbf{74.4}$ &              $52.4$ &  \underline{$53.7$} &&     $\mathbf{5.5}$ &   \underline{$7.6$} &               $7.7$ \\
\texttt{text-davinci-002}          &   \underline{$4.9$} &   \underline{$4.9$} &      $\mathbf{8.5}$ &&  \underline{$9.7$} &               $9.8$ &      $\mathbf{9.6}$ \\
\texttt{text-davinci-003}          &     $\mathbf{20.7}$ &  \underline{$18.3$} &               $7.3$ &&  \underline{$9.2$} &      $\mathbf{9.0}$ &               $9.6$ \\

\midrule
\multicolumn{7}{c}{\textbf{Frequency of Best-performing Format} $\uparrow$} \\
Open-source & $\mathbf{5}$ & \underline{$4$} & $3$ && $\mathbf{6}$ & \underline{$1$} & \underline{$1$} \\

Closed-source & $\mathbf{7}$ & \underline{$1$} & \underline{$1$} && $\mathbf{6}$ & \underline{$2$} & $1$ \\

\textbf{Overall} & $\mathbf{12}$ & \underline{5} & $4$ && $\mathbf{12}$ &  \underline{$3$} & $2$ \\

\bottomrule
\end{tabular}
\end{adjustbox}
\end{minipage}
\end{table*}

\subsection{\approach Gets More Done with Fewer Interactions}
\label{sec:control-data-flow}

In this section, we investigate whether LLM agents can benefit from the control and data flow of code on problems that require complex patterns of tool use.

\noindent \textbf{\evalname.}
As shown in \tref{tab:tool_bench_comparison}, to the best of our knowledge, no existing tool-use benchmarks contain complex tasks requiring the composition of multiple tools while supporting evaluating different action formats. Hence, we curate a benchmark \evalname to fill this gap, which evaluates LLMs' capabilities in solving complex tasks that typically require \textbf{m}ultiple calls to \textbf{m}ultiple tools in \textbf{m}ulti-turn interactions.
It contains 82 human-curated instances, spanning tasks including web browsing, finance, travel itinerary planning, science, and information processing. 
Each domain is accompanied by a unique set of manually crafted tools.
We intentionally keep the prompt simple (examples in \sref{sec:zeroshot_act_prompt}) and avoid providing any demonstration to test the LLM's zero-shot ability to use tools, similar to how a novice user without knowledge of few-shot prompting would use the model.
%

\noindent \textbf{Setup.}
We allow the model to generate fully functional Python code that enables control and data flow (e.g., if-statement, for-loop). We follow the action format for JSON and text described in \tref{tab:apibank-action-example}.
Within each turn, the model can either emit an \textit{action} or propose an \textit{answer} to be verified by an exact match with the ground-truth solution.
The interaction will terminate when a maximum of 10 interaction turns are reached or a correct solution has been submitted, similar to \cite{wang2023mint}.

\noindent \textbf{Metric.} We measure the \textit{success rate} by calculating the percentage of the model proposed answers that match the ground-truth solutions. We also include the \textit{avg. turns} metric: the average number of turns on all evaluated instances.

\noindent \textbf{Quantitative Results on \evalname.}
We include full results in \tref{tab:zeroshot_act_results} and a subset of results for visualization in \fref{fig:zeroshot_acteval}.
\approach generally has a higher task success rate (12 out of 17 evaluated LLMs), similar to the trend in \sref{sec:data-factor-format}. Moreover, using \approach requires a lower average number of turns (12 out of 17 evaluated LLMs).
For example, the best model \texttt{gpt-4-1106-preview} achieves a $20.7$\% absolute improvement compared to the next best action format (text) while requiring $2.1$ fewer interaction turns on average.
However, there is still a significant gap in terms of absolute \approach performance between open- and closed-source LLMs as the best open-source model achieving 13.4\% while the best closed-source model \texttt{gpt-4-1106-preview} 74.4\%. This is potentially due to open-source models' weak task-solving capability and inability to follow complex instructions without demonstration, suggesting an urgent need to improve open-source LLMs for practical, real-world tasks under the zero-shot setting.

\begin{figure*}[!ht]
    \centering
    \includegraphics[width=\textwidth]{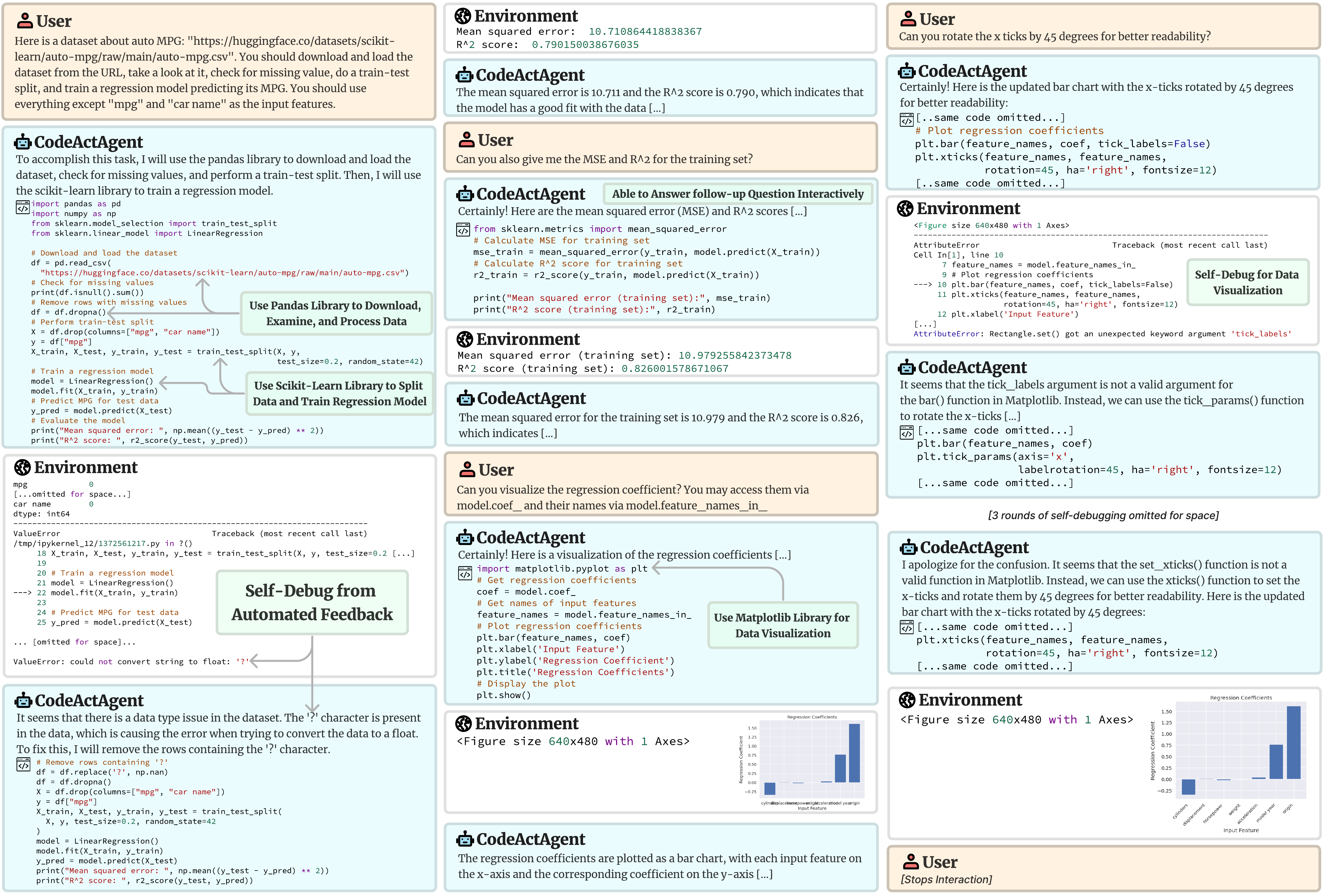}
    \vspace{-20pt}
    \caption{Example multi-turn interaction with Python packages using \modelname (Mistral-7b). No in-context demonstrations are provided to the model. Some messages are omitted for space.
    See \url{https://chat.xwang.dev/r/Vqn108G} for complete interaction.
    }
    \label{fig:qualitative_example}
\end{figure*}

\subsection{\approach Benefits from Multi-turn Interactions and Existing Software Packages}
\label{sec:multiturn_software_benefit}

In \fref{fig:qualitative_example}, we show how an LLM agent can integrate with Python (i.e., \modelname we trained in \sref{sec:llm_agent_evaluation}) and use existing software to perform complex tasks in multi-turn interactions.
Thanks to its extensive knowledge of Python learned during pre-training, the LLM agent can automatically import the correct \textit{Python libraries} to solve tasks without requiring user-provided tools or demonstrations.
As illustrated in \fref{fig:qualitative_example}, \modelname can use Pandas to download and process tabular data, use Scikit-Learn for machine learning train-test data split and regression model training, and use Matplotlib for data visualization.
Furthermore, using the interactive Python interpreter for code execution allows automated error messages that help the LLM agent `self-debug' their actions in a \textit{multi-turn interaction} and eventually complete the human user's request correctly.

\looseness=-1
\section{Empowering Open-source LLM Agent to be Better at \approach}
\label{sec:data_and_evaluation}
\label{sec:llm_agent_framework_capability}

The promising results achieved by \approach motivate us to build an open-source LLM agent that can both interact with environments through \approach and communicate with humans using language.
To improve open-source LLMs' \approach capability, in \sref{sec:agent_env_data}, we introduce \dataname, an instruction finetuning dataset that contains agent-environment interaction trajectories.
We discuss data selection procedures in \sref{sec:agent_env_data_self_improve} to promote improvement from interaction behavior.
Additionally, we show that \approach can be used together with existing agent-user conversation data (\sref{sec:agent_user_data}) to balance the dialog capability of the resulting LLM.
Our model \modelname, finetuned from LLaMA-2 \citep{touvron2023llama} and Mistral-7B \citep{jiang2023mistral} on a mixture of \dataname and general conversations, improves \approach performances without hurting LLM's general performance on a diverse suite of tasks (\sref{sec:llm_agent_evaluation}).

\subsection{\dataname: Agent-Environment Interactions}
\label{sec:agent_env_data}

We consider four main use cases in agent-environment interaction and repurpose five existing datasets across different domains to generate trajectories:

\begin{itemize}[noitemsep,topsep=0pt,parsep=0pt,partopsep=0pt,leftmargin=*]
    \item \textbf{Information Seeking:} We use a training subset of HotpotQA \citep{yang2018hotpotqa} to generate information-seeking trajectories, where LLMs use the \texttt{wikipedia\_search} API (provided as a Python function) to search for information to answer questions.
    
    \item \textbf{Software Package (Tool) Usage:} We use the training set of code generation problems in APPS \citep{hendrycks2021measuring} and math problems in MATH \citep{hendrycks2021math}. The code generation tasks already involve importing packages and/or creating new tools by defining a new Python function. For MATH, we provide an in-context demonstration of importing Python packages (e.g., \texttt{sympy} for symbolic math) for problem-solving.
    
    \item \textbf{External Memory:} We repurpose the training subset of WikiTableQuestion \citep{pasupat2015compositional} and tweak it into two variants of tabular reasoning tasks that require accessing external memory: (1) SQL-based, requiring the LLM to interact with an SQL database through sqlite3 package to answer the question via SQL execution; (2) Pandas-based, requiring the model to interact with pandas tables to perform data operations (e.g., select, filter). Examples of instructions can be found in \sref{sec:interaction_data_gen_prompt_tabular}.

    \item \textbf{Robot Planning:} We use ALFWorld \citep{shridhar2020alfworld}, a text-only embodied environment simulator, to generate trajectories that use robot-control APIs (repurposed as Python function) to complete household tasks. Following MINT \citep{wang2023mint}, we provide an in-context demonstration to encourage the use of for-loop and if-statement code blocks to automate repetitive operations (e.g., searching for items by visiting different locations).
    
\end{itemize}

\noindent \textbf{Data Down-sampling.}
We down-sample each dataset by keeping only the most challenging instances, aiming to make trajectory generation more efficient and cost-effective. Furthermore, it also helps remove simple instances that existing LLMs can already solve.
The statistics of the filtered dataset can be found in \tref{tab:data_training_instances}. Please refer to \sref{sec:how_to_downsample_dataset} for details about the down-sample process.

\noindent \textbf{Repurpose Data for Multi-turn Interaction.}
Some datasets (APPS, MATH, WikiTableQuestions) are initially single-turn problems that expect \textit{one} solution per instruction, whereas, in a realistic agent use case, we often require multi-turn interaction to complete each task (\fref{fig:illustrative_example} top).
Following MINT \citep{wang2023mint}, we repurpose single-turn problems into multi-turn ones by allowing LLM to interact with the environment for multiple turns before it decides to submit one solution for evaluation.
Specifically for code generation problems, we provide an in-context example to guide LLMs to test their solution on provided test cases before they submit the solution.
Metrics from the original data will evaluate the submitted solution to determine its correctness. We include examples in \sref{sec:interaction_data_gen_prompt}.

\noindent \textbf{Trajectory Generation.}
We use MINT's evaluation framework \citep{wang2023mint} to generate interaction trajectories for the aforementioned datasets and determine the correctness of each trajectory.
We run gpt-3.5-turbo-0613 from OpenAI, claude-1-instant and claude-2 from Anthropic on down-sampled data, except code generation, which we use a longer-context version of GPT-3.5 (gpt-3.5-turbo-0613-16k) due to the long-context requirement of the self-debugging process.
On a subset of problems that none of these models can solve, we use gpt-4-0613 to generate trajectories.

\noindent \textbf{Enhancing Agent's Capabilities of Improving from Interaction.}
\label{sec:agent_env_data_self_improve}
We select a high-quality subset of all the generated trajectories from \dataname to promote the agent's ability to improve the next action based on prior observations (e.g., self-debugging from code execution error message, a planning capability in \fref{fig:llm_agent_framework}).
To achieve this, we selectively preserve those trajectories wherein the model initially encounters errors but rectifies these inaccuracies in later interactions.
For these instances, the LLM typically engages in self-reflection following the initial error, thereby proactively enhancing its future actions. 
Other filtering details are discussed in \sref{sec:data_filter_heuristic}.
On all trajectories generated, we keep 411 trajectories from gpt-4-0613 and 6728 trajectories from gpt-3.5 and claude. 
The statistics of the resulting dataset \dataname are shown in \tref{tab:data_training_mixture_stats}.

\begin{table*}[!ht]
\centering
\caption{Statistics of our training mixture and comparison with prior work. Please refer to \sref{sec:agent_env_data} for details about \dataname and general conversation data. Token statistics are computed using Llama-2 tokenizer.
}
\vspace{-9pt}
\resizebox{\textwidth}{!}{
\begin{tabular}{lrrrrr}
\toprule
\textbf{Data Mixture} & \textbf{Data Type} & \textbf{Data Name} & \textbf{\# of Data Instances} &  \textbf{\# of Total Tokens} & \textbf{Avg. Tokens Per Instance} \\
\midrule
\multirow{2}{*}{\textbf{Prior Work}}& - & FireAct \citep{chen2023fireact} &   $2,063$ &    $542,176$ &   $262.81$ \\
 &  - & AgentInstruct \citep{zeng2023agenttuning} &    $1,866$ &   $2,517,785$ &  $1349.30$ \\
\midrule

\multirow{6}{*}{\textbf{\dataname} (Ours)}
& Information Seeking & HotpotQA \citep{yang2018hotpotqa} &      $1,664$ &   $2,472,227$ &  $1485.71$ \\
& Software Packages (Tool) & MATH (Math, \citep{hendrycks2021math}) &      $1,732$ &   $1,719,467$ &   $992.76$ \\
& Software Packages (Tool) & APPS (Code, \citep{hendrycks2021measuring}) &        $647$ &   $1,235,472$ &  $1909.54$ \\
& External Memory & WikiTableQuestion \citep{pasupat2015compositional} &      $1,065$ &   $1,316,246$ &  $1235.91$ \\
& Robot Planning & ALFWorld \citep{shridhar2020alfworld} &      $2,031$ &   $3,838,269$ &  $1889.84$ \\
\cmidrule{2-6}
& \multicolumn{2}{r}{\textbf{Total}} &      $\mathbf{7,139}$ &  $\mathbf{10,581,681}$ &  $\mathbf{1482.24}$ \\

\midrule

\multirow{5}{*}{\textbf{General Conversation}}
& Single-Turn Reasoning & OpenOrca (Sub-sampled, \citep{OpenOrca}) &     $50,000$ &  $14,034,152$ &   $280.68$ \\
& Multi-Turn Conversations & ShareGPT (Sub-sampled, \citep{sharegpt_dataset}) &     $10,000$ &  $17,933,861$ &  $1793.39$ \\
& Multi-Turn Conversations & ShareGPT (GPT-4, \citep{sharegpt4_dataset}) &      $4,583$ &  $18,195,878$ &  $3970.30$ \\
& Multi-turn Reasoning & CapyBara \citep{capybara_dataset} &  $4,647$ &   $4,982,435$ &  $1072.18$ \\
\cmidrule{2-6}
& \multicolumn{2}{r}{\textbf{Total}} &     $\mathbf{69,230}$ &  $\mathbf{55,146,326}$ &   $\mathbf{796.57}$ \\

\bottomrule
\end{tabular}
}
\label{tab:data_training_mixture_stats}
\end{table*}

\noindent \textbf{Comparing \dataname with Prior Work.}
Compared with prior work AgentInstruct \citep{zeng2023agenttuning} and FireAct \citep{chen2023fireact} that mainly focus using text as action, \dataname results in models that are more practical in real-world implementation, as such models using \approach can directly interact with Python interpreters and open-source toolkits (\fref{fig:qualitative_example}), reducing the development effort for action parsing and tool creations.
\dataname is systematically constructed following the general agent framework (\fref{fig:llm_agent_framework}).
It covers diverse domains (e.g., compared to FireAct that only considers QA-task and search API), contains quality data (e.g., promotes agent's capability of self-debug) and of larger size (3.8x / 3.5x more data trajectories and 5x / 19x more tokens compared to AgentInstruct / FireAct respectively in \tref{tab:data_training_mixture_stats}). 
As we empirically show in \tref{tab:model_benchmark_results}, the resulting model (same backbone) of \dataname achieves 24\% and 119\% relative improvement compared to AgentInstruct and FireAct.

\noindent \textbf{\dataname Can Be Used With Existing Agent-User Conversation Data.}
\label{sec:agent_user_data}
We use a sub-sampled set of OpenOrca \citep{OpenOrca} that focuses on single-turn chain-of-thought (CoT) reasoning, ShareGPT \citep{sharegpt_dataset, sharegpt4_dataset} from two sources that contain multi-turn conversations between human and LLM, and CapyBara \citep{capybara_dataset} that focuses on reasoning in multi-turn conversations.
Statistics and down-sampling details can be found in \tref{tab:data_training_mixture_stats} and \sref{sec:general_data_downsample}.
%

{
\begin{table*}[!t]
\centering
\caption{
Evaluation results for \modelname. The best results among all open-source LLMs are \textbf{bolded}, and the second-best results are \underline{underlined}.
ID and OD stand for in-domain and out-of-domain evaluation correspondingly.
Overall averaged performance normalizes the MT-Bench score to be consistent with other tasks and excludes in-domain tasks for fair comparison.
}
\vspace{-10pt}
\begin{threeparttable}
\begin{adjustbox}{max width=\textwidth}
\begin{tabular}{@{} lc rrr m{0.01em} rr m{0.01em} rrrr m{0.01em} r@{}}
\toprule
& & \multicolumn{6}{c}{\textbf{Agent Tasks}}                                                         && \multicolumn{4}{c}{\textbf{Generic Tasks}} && \textbf{Overall} \\
 \cmidrule{3-8}
& & \multicolumn{3}{c}{\textbf{Code as Action}} && \multicolumn{2}{c}{\textbf{Text as Action (OD)}} && \multicolumn{4}{c}{\textbf{(OD)}} && \textbf{Average}\\
  \cmidrule{3-5}
  \cmidrule{7-8}
  \cmidrule{10-14}
 Model & Size & MINT (ID) & MINT (OD) & \evalname (OD) && Miniwob++ & SciWorld && MMLU & HumanEval & GSM8K & MTBench & \\
\midrule

\multicolumn{15}{c}{\textit{Open-source LLMs (LLaMA-2-based)}} \\

Llama2 Base & 7B & -$^*$ & -$^*$ & -$^*$ &  & -$^*$ & -$^*$ &  & $45.3$ & $12.8$ & $14.6$ & -$^*$ &  & -$^*$ \\
Llama2 Chat & 7B & $3.2$ & $11.0$ & \underline{$0.0$} &  & $0.0$ & $5.9$ &  & $48.0$ & $13.9$ & $27.7$ & $6.3$ &  & $21.1$ \\
FireAct \citep{chen2023fireact} & 7B & $0.0$ & $0.3$ & \underline{$0.0$} &  & $0.0$ & $6.8$ &  & $44.1$ & $3.5$ & $12.4$ & $4.5$ &  & $14.0$ \\
AgentLM \citep{zeng2023agenttuning} & 7B & $8.7$ & $6.1$ & \underline{$0.0$} &  & \underline{$28.9$} & $13.7$ &  & $48.7$ & $15.4$ & $24.6$ & $6.1$ &  & $24.8$ \\
\modelname (LLaMA-2) & 7B & \underline{$51.3$} & \underline{$20.4$} & \underline{$0.0$} &  & $25.5$ & $\mathbf{17.6}$ &  & $50.6$ & $18.1$ & $38.3$ & \underline{$7.5$} &  & \underline{$30.7$} \\

\midrule
\multicolumn{15}{c}{\textit{Open-source LLMs (Mistral-based)}} \\

Mistral Base & 7B & -$^*$ & -$^*$ & -$^*$ &  & -$^*$ & -$^*$ &  & $\mathbf{60.1}$ & \underline{$30.5$} & \underline{$52.1$} & -$^*$ &  & -$^*$ \\
Mistral Instruct & 7B & $18.8$ & $9.7$ & \underline{$0.0$} &  & $0.5$ & $4.0$ &  & $53.8$ & $29.3$ & $43.3$ & $6.4$ &  & $25.6$ \\
\modelname (Mistral) & 7B & $\mathbf{57.4}$ & $\mathbf{32.4}$ & $\mathbf{12.2}$ &  & $\mathbf{46.2}$ & \underline{$15.9$} &  & \underline{$59.1$} & $\mathbf{34.7}$ & $\mathbf{58.0}$ & $\mathbf{8.2}$ &  & $\mathbf{42.5}$ \\

\midrule
\multicolumn{15}{c}{\textit{Closed-source LLMs}} \\

gpt-3.5-turbo-0613 & - & $33.9$ & $38.2$ & $51.2$ &  & $66.7$ & $21.2$ &  & $70.0$ & $48.1$ & $57.1$ & $7.9$ &  & $54.0$ \\
gpt-4-0613 & - & $68.6$ & $70.2$ & $67.1$ &  & $69.4$ & $36.4$ &  & $86.4$ & $67.0$ & $87.1$ & $9.0$ &  & $71.7$ \\

\bottomrule
\end{tabular}
\end{adjustbox}
{
\small
\begin{tablenotes}
    \item[*] Some results are only available with instruction-tuned models.
\end{tablenotes}
}
\end{threeparttable}
\label{tab:model_benchmark_results}
\vspace{-15pt}
\end{table*}
}

\subsection{\modelname}
\label{sec:llm_agent_evaluation}

We fine-tune Llama-2 7B \citep{touvron2023llama} and Mistral 7B \citep{jiang2023mistral} on a mixture of \dataname and general conversations (\tref{tab:data_training_mixture_stats}) to obtain \modelname.

\noindent \textbf{Training Setup.} We perform full-parameter supervised fine-tuning with a sequence length of 4,096 tokens for Llama-2 and 16,384 for Mistral. Please refer to \sref{sec:model_training_details} for more details.

\noindent \textbf{Evaluation Setup.} 
We use MINT \citep{wang2023mint} to evaluate LLMs with \approach on a diverse range of agent tasks.
\modelname has some training domains overlapping with MINT's evaluation (i.e., MINT includes ALFWorld and MATH), hence we report separate numbers for MINT's in- and out-of-domain performance.
Unless otherwise specified, we measure MINT tasks' success rates with interaction turn $k=5$.
We also evaluate out-of-domain agent tasks using text actions from MiniWob++ (computer tasks, \cite{kim2023language}) and ScienceWorld (text-based simulator for elementary science curriculum, \cite{Wang2022ScienceWorldIY}) to test whether \modelname can generalize to different action formats.
Finally, we include a suite of general LLM evaluation tasks to assess general capability: MMLU \citep{hendrycks2020measuring} for knowledge-based QA, HumanEval \citep{chen2021evaluating} for single-turn code-generation, GSM8K \citep{cobbe2021training} for single-turn tool-free math reasoning, and MTBench \citep{zheng2023judging} for instruction-following.

\noindent \textbf{\modelname Excels in \approach Task.}
As shown in \tref{tab:model_benchmark_results}, \modelname (both variants) perform better than all evaluated open-source LLMs on both the in- and out-of-domain subsets of MINT.
On \evalname, we find \modelname (Mistral) outperforms open-source LLMs of similar size (7B and 13B) and even reaches similar performance to those 70B models (\tref{tab:zeroshot_act_results}).
Surprisingly, no improvement is observed for the Llama-2 variant. We discuss potential reasons in \sref{sec:llama_anomaly}.

\noindent \textbf{\modelname Generalizes to Text Action.}
When evaluated on out-of-domain text actions, \modelname (LLaMA2, 7B), which has never been optimized for text action, achieves comparable performance to AgentLM-7B \citep{zeng2023agenttuning} which has explicit tuning for text actions.

\noindent \textbf{\modelname Maintains or Improves the Performance on General LLM Tasks.}
In \tref{tab:model_benchmark_results}, we find that \modelname (both variants) performs better on generic LLM tasks we tested, except for a slight degradation on MMLU for \modelname (Mistral, 7B).

\noindent \textbf{Ablation Study.}
\tref{tab:ablation_study_results} presents ablation experiments to determine the importance of \dataname and general conversations.
Both \dataname and general conversations contribute to agent tasks, while general conversations are essential to maintain performance on general tasks.
%

\looseness=-1
\section{Related Work}
\subsection{Action Module in LLM Agents}
As detailed in \cite{wang2023survey}, LLM-based autonomous agents are typically structured around four components: customized profiles~\citep{park2023generative, qian2023communicative}, long-term memory capabilities~\citep{zhu2023ghost, fischer2023reflective}, reasoning and planning algorithms~\citep{wei2022chain, chen2023measuring}, and, most crucially, action modules. 
The action modules are key to facilitating LLM agents to effectively interact with external entities, including humans~\citep{lee2022coauthor} and tools~\citep{qin2023tool} in the environment~\citep{wang2023mint,yang2024intercode}. 
In this study, we address the critical problem of standardizing the action space for LLM agents. 
We further discuss the difference between \approach and the line of work that uses code generation for problem-solving in \sref{sec:prior_code_action_work}.
We notice a concurrent study TaskWeaver~\citep{qiao2023taskweaver} similarly endorses the use of code. We discuss the principal distinctions in \sref{sec:taskweaver}.

\subsection{Improving LLM Agents}

Two primary methods for enhancing LLM agents are prompt engineering and instruction tuning, as surveyed by \cite{wang2023survey}.
For \textit{prompt engineering}~\citep{liu2023pre}, numerous strategies have been introduced to improve the chain-of-thought reasoning~\citep{wei2022chain}, including self-consistency-based reasoning~\citep{wang2022self, chen2023measuring} and tree-based approaches~\citep{yao2023tree}. 
Moreover, LLMs can be strategically prompted to reflect on previous plans~\citep{yao2023retroformer, wang2023describe, zhang2023prefer}, enabling them to refine initial actions through trial and error.
Contrast to prompt engineering, \textit{instruction tuning} intrinsically enhances LLMs~\citep{chung2022scaling}, particularly in their agent capabilities~\citep{zeng2023agenttuning, chen2023fireact}. 
For effective training, human annotators can curate expert demonstrations for specific agent tasks, such as web browsing \citep{yao2022webshop, nakano2021webgpt}. 
%
%
To minimize human annotation efforts, prior work creates synthetic datasets using stronger LLMs to distill agent capabilities into local models, focusing on tool usage~\citep{Qin2023ToolLLMFL}, interaction~\citep{chen2023dress}, and social skills~\citep{liu2023training}.
\dataname aligns with the latter approach and creates datasets using stronger LLMs.
\section{Conclusions}
This work introduces \approach that employs executable Python code for the LLM agent's action, which is advantageous over using text or JSON action, especially in complex scenarios.
We collect \approach-focused multi-turn interaction trajectories \dataname for instruction tuning, and train \modelname that is specially designed for seamless integration with Python and can execute sophisticated tasks (e.g., model training) leveraging existing Python packages and autonomously rectifying errors through self-debugging.

\section*{Acknowledgement}
We thank the anonymous reviewers for their suggestions and comments.
This research is based upon work supported by U.S. DARPA ECOLE Program No. HR00112390060 and U.S. DARPA ITM Program No. FA8650-23-C-7316 and KAIROS Program No. FA8750-19-2-1004. The views and conclusions contained herein are those of the authors and should not be interpreted as necessarily representing the official policies, either expressed or implied, of DARPA, or the U.S. Government. The U.S. Government is authorized to reproduce and distribute reprints for governmental purposes notwithstanding any copyright annotation therein.
This work used the Delta system at the National Center for Supercomputing Applications through allocation CIS230256 from the Advanced Cyberinfrastructure Coordination Ecosystem: Services \& Support (ACCESS, \citealt{boerner2023access}) program, which is supported by National Science Foundation grants \#2138259, \#2138286, \#2138307, \#2137603, and \#2138296.

\section*{Impact Statement}
This paper presents work whose goal is to advance LLM-based autonomous agents that can communicate with humans through natural language and assist human users by performing tasks in environments on behalf of humans.
In this section, we discuss potential societal consequences, limitations, and future work related to our work and its goal.

\modelname is an initial prototype of an autonomous agent and still has several practical limitations. For example, it may suffer from hallucination commonly seen in LLMs (e.g., imagine the content of a variable without actually printing it out), suggesting the need for subsequent alignment \citep{ouyang2022training} for further improvements.

Despite being a prototype, \modelname has already demonstrated limited self-improving capability (e.g., self-debug error messages to improve its action) and the ability to interact with environments.
Future work may build upon \modelname to develop better agents by having them perform extensive interactions within a given environment and iteratively bootstrap their self-improving capability to learn to improve from past mistakes.
More powerful agents, as results of such algorithms, are potentially beneficial for solving a wide range of real-world problems  (e.g., theorem proving, drug discovery).
As extensively discussed in \cite{eloundou2023gpts}, a fully autonomous agent may transform the current landscape of the labor market and impact the jobs of existing workers.

Furthermore, since \approach directly grants access for the agent to freely execute code in a sandbox environment, in the worst scenario (e.g., in Sci-Fi movies), such an agent may potentially break free of the sandbox restriction and cause harm to the world through cyber-attack, highlighting the need for future work to design better safety mechanism to safeguard autonomous agents \citep{tang2024prioritizing}.

\bibliography{custom}

\begin{thebibliography}{75}
\providecommand{\natexlab}[1]{#1}
\providecommand{\url}[1]{\texttt{#1}}
\expandafter\ifx\csname urlstyle\endcsname\relax
  \providecommand{\doi}[1]{doi: #1}\else
  \providecommand{\doi}{doi: \begingroup \urlstyle{rm}\Url}\fi

\bibitem[Ahn et~al.(2022)Ahn, Brohan, Brown, Chebotar, Cortes, David, Finn, Fu, Gopalakrishnan, Hausman, Herzog, Ho, Hsu, Ibarz, Ichter, Irpan, Jang, Ruano, Jeffrey, Jesmonth, Joshi, Julian, Kalashnikov, Kuang, Lee, Levine, Lu, Luu, Parada, Pastor, Quiambao, Rao, Rettinghouse, Reyes, Sermanet, Sievers, Tan, Toshev, Vanhoucke, Xia, Xiao, Xu, Xu, Yan, and Zeng]{saycan2022arxiv}
Ahn, M., Brohan, A., Brown, N., Chebotar, Y., Cortes, O., David, B., Finn, C., Fu, C., Gopalakrishnan, K., Hausman, K., Herzog, A., Ho, D., Hsu, J., Ibarz, J., Ichter, B., Irpan, A., Jang, E., Ruano, R.~J., Jeffrey, K., Jesmonth, S., Joshi, N., Julian, R., Kalashnikov, D., Kuang, Y., Lee, K.-H., Levine, S., Lu, Y., Luu, L., Parada, C., Pastor, P., Quiambao, J., Rao, K., Rettinghouse, J., Reyes, D., Sermanet, P., Sievers, N., Tan, C., Toshev, A., Vanhoucke, V., Xia, F., Xiao, T., Xu, P., Xu, S., Yan, M., and Zeng, A.
\newblock Do as i can and not as i say: Grounding language in robotic affordances.
\newblock In \emph{arXiv preprint arXiv:2204.01691}, 2022.

\bibitem[Anonymous(2023)]{sharegpt_dataset}
Anonymous.
\newblock Sharegpt dataset.
\newblock \url{https://hf.co/datasets/anon8231489123/ShareGPT_Vicuna_unfiltered/blob/main/ShareGPT_V3_unfiltered_cleaned_split_no_imsorry.json}, 2023.
\newblock A dataset containing multi-turn conversations between human and LLM assistant.

\bibitem[Boerner et~al.(2023)Boerner, Deems, Furlani, Knuth, and Towns]{boerner2023access}
Boerner, T.~J., Deems, S., Furlani, T.~R., Knuth, S.~L., and Towns, J.
\newblock Access: Advancing innovation: Nsf’s advanced cyberinfrastructure coordination ecosystem: Services \& support.
\newblock In \emph{Practice and Experience in Advanced Research Computing}, pp.\  173--176. 2023.

\bibitem[Bran et~al.(2023)Bran, Cox, White, and Schwaller]{bran2023chemcrow}
Bran, A.~M., Cox, S., White, A.~D., and Schwaller, P.
\newblock Chemcrow: Augmenting large-language models with chemistry tools.
\newblock \emph{arXiv preprint arXiv:2304.05376}, 2023.

\bibitem[Cano et~al.(2023)Cano, Pagliardini, Köpf, Matoba, Mohtashami, Wang, Fan, Marmet, Bayazit, Krawczuk, Chen, Salvi, Bosselut, and Jaggi]{epfmgtrn}
Cano, A.~H., Pagliardini, M., Köpf, A., Matoba, K., Mohtashami, A., Wang, X., Fan, O.~S., Marmet, A., Bayazit, D., Krawczuk, I., Chen, Z., Salvi, F., Bosselut, A., and Jaggi, M.
\newblock epfllm megatron-llm, 2023.
\newblock URL \url{https://github.com/epfLLM/Megatron-LLM}.

\bibitem[Chase(2022)]{Chase_LangChain_2022}
Chase, H.
\newblock {LangChain}, October 2022.
\newblock URL \url{https://github.com/langchain-ai/langchain}.

\bibitem[Chen et~al.(2023{\natexlab{a}})Chen, Shu, Shareghi, Collier, Narasimhan, and Yao]{chen2023fireact}
Chen, B., Shu, C., Shareghi, E., Collier, N., Narasimhan, K., and Yao, S.
\newblock Fireact: Toward language agent fine-tuning.
\newblock \emph{arXiv preprint arXiv:2310.05915}, 2023{\natexlab{a}}.

\bibitem[Chen et~al.(2021)Chen, Tworek, Jun, Yuan, Pinto, Kaplan, Edwards, Burda, Joseph, Brockman, et~al.]{chen2021evaluating}
Chen, M., Tworek, J., Jun, H., Yuan, Q., Pinto, H. P. d.~O., Kaplan, J., Edwards, H., Burda, Y., Joseph, N., Brockman, G., et~al.
\newblock Evaluating large language models trained on code.
\newblock \emph{arXiv preprint arXiv:2107.03374}, 2021.

\bibitem[Chen et~al.(2023{\natexlab{b}})Chen, Lin, Sch{\"a}rli, and Zhou]{chen2023teaching}
Chen, X., Lin, M., Sch{\"a}rli, N., and Zhou, D.
\newblock Teaching large language models to self-debug.
\newblock \emph{arXiv preprint arXiv:2304.05128}, 2023{\natexlab{b}}.

\bibitem[Chen et~al.(2023{\natexlab{c}})Chen, Sikka, Cogswell, Ji, and Divakaran]{chen2023dress}
Chen, Y., Sikka, K., Cogswell, M., Ji, H., and Divakaran, A.
\newblock Dress: Instructing large vision-language models to align and interact with humans via natural language feedback.
\newblock \emph{arXiv preprint arXiv:2311.10081}, 2023{\natexlab{c}}.

\bibitem[Chen et~al.(2023{\natexlab{d}})Chen, Sikka, Cogswell, Ji, and Divakaran]{chen2023measuring}
Chen, Y., Sikka, K., Cogswell, M., Ji, H., and Divakaran, A.
\newblock Measuring and improving chain-of-thought reasoning in vision-language models.
\newblock \emph{arXiv preprint arXiv:2309.04461}, 2023{\natexlab{d}}.

\bibitem[Chung et~al.(2022)Chung, Hou, Longpre, Zoph, Tay, Fedus, Li, Wang, Dehghani, Brahma, et~al.]{chung2022scaling}
Chung, H.~W., Hou, L., Longpre, S., Zoph, B., Tay, Y., Fedus, W., Li, Y., Wang, X., Dehghani, M., Brahma, S., et~al.
\newblock Scaling instruction-finetuned language models.
\newblock \emph{arXiv preprint arXiv:2210.11416}, 2022.

\bibitem[Cobbe et~al.(2021)Cobbe, Kosaraju, Bavarian, Chen, Jun, Kaiser, Plappert, Tworek, Hilton, Nakano, et~al.]{cobbe2021training}
Cobbe, K., Kosaraju, V., Bavarian, M., Chen, M., Jun, H., Kaiser, L., Plappert, M., Tworek, J., Hilton, J., Nakano, R., et~al.
\newblock Training verifiers to solve math word problems.
\newblock \emph{arXiv preprint arXiv:2110.14168}, 2021.

\bibitem[Eloundou et~al.(2023)Eloundou, Manning, Mishkin, and Rock]{eloundou2023gpts}
Eloundou, T., Manning, S., Mishkin, P., and Rock, D.
\newblock Gpts are gpts: An early look at the labor market impact potential of large language models.
\newblock \emph{arXiv preprint arXiv:2303.10130}, 2023.

\bibitem[Fischer(2023)]{fischer2023reflective}
Fischer, K.~A.
\newblock Reflective linguistic programming (rlp): A stepping stone in socially-aware agi (socialagi).
\newblock \emph{arXiv preprint arXiv:2305.12647}, 2023.

\bibitem[Gao et~al.(2023)Gao, Madaan, Zhou, Alon, Liu, Yang, Callan, and Neubig]{gao2023pal}
Gao, L., Madaan, A., Zhou, S., Alon, U., Liu, P., Yang, Y., Callan, J., and Neubig, G.
\newblock Pal: Program-aided language models.
\newblock In \emph{International Conference on Machine Learning}, pp.\  10764--10799. PMLR, 2023.

\bibitem[Hendrycks et~al.(2020)Hendrycks, Burns, Basart, Zou, Mazeika, Song, and Steinhardt]{hendrycks2020measuring}
Hendrycks, D., Burns, C., Basart, S., Zou, A., Mazeika, M., Song, D., and Steinhardt, J.
\newblock Measuring massive multitask language understanding.
\newblock In \emph{International Conference on Learning Representations}, 2020.

\bibitem[Hendrycks et~al.(2021{\natexlab{a}})Hendrycks, Basart, Kadavath, Mazeika, Arora, Guo, Burns, Puranik, He, Song, et~al.]{hendrycks2021measuring}
Hendrycks, D., Basart, S., Kadavath, S., Mazeika, M., Arora, A., Guo, E., Burns, C., Puranik, S., He, H., Song, D., et~al.
\newblock Measuring coding challenge competence with apps.
\newblock In \emph{Thirty-fifth Conference on Neural Information Processing Systems Datasets and Benchmarks Track (Round 2)}, 2021{\natexlab{a}}.

\bibitem[Hendrycks et~al.(2021{\natexlab{b}})Hendrycks, Burns, Kadavath, Arora, Basart, Tang, Song, and Steinhardt]{hendrycks2021math}
Hendrycks, D., Burns, C., Kadavath, S., Arora, A., Basart, S., Tang, E., Song, D., and Steinhardt, J.
\newblock Measuring mathematical problem solving with the math dataset.
\newblock In \emph{Thirty-fifth Conference on Neural Information Processing Systems Datasets and Benchmarks Track (Round 2)}, 2021{\natexlab{b}}.

\bibitem[Hong et~al.(2023)Hong, Zheng, Chen, Cheng, Wang, Zhang, Wang, Yau, Lin, Zhou, et~al.]{hong2023metagpt}
Hong, S., Zheng, X., Chen, J., Cheng, Y., Wang, J., Zhang, C., Wang, Z., Yau, S. K.~S., Lin, Z., Zhou, L., et~al.
\newblock Metagpt: Meta programming for multi-agent collaborative framework.
\newblock \emph{arXiv preprint arXiv:2308.00352}, 2023.

\bibitem[Hong et~al.(2024)Hong, Lin, Liu, Liu, Wu, Li, Chen, Zhang, Wang, Zhang, Zhang, Yang, Zhuge, Guo, Zhou, Tao, Wang, Tang, Lu, Zheng, Liang, Fei, Cheng, Xu, and Wu]{hong2024data}
Hong, S., Lin, Y., Liu, B., Liu, B., Wu, B., Li, D., Chen, J., Zhang, J., Wang, J., Zhang, L., Zhang, L., Yang, M., Zhuge, M., Guo, T., Zhou, T., Tao, W., Wang, W., Tang, X., Lu, X., Zheng, X., Liang, X., Fei, Y., Cheng, Y., Xu, Z., and Wu, C.
\newblock Data interpreter: An llm agent for data science, 2024.

\bibitem[Huang et~al.(2023)Huang, Wang, Zhang, Li, Wu, and Fei-Fei]{huang2023voxposer}
Huang, W., Wang, C., Zhang, R., Li, Y., Wu, J., and Fei-Fei, L.
\newblock Voxposer: Composable 3d value maps for robotic manipulation with language models.
\newblock \emph{arXiv preprint arXiv:2307.05973}, 2023.

\bibitem[Jiang et~al.(2023)Jiang, Sablayrolles, Mensch, Bamford, Chaplot, Casas, Bressand, Lengyel, Lample, Saulnier, et~al.]{jiang2023mistral}
Jiang, A.~Q., Sablayrolles, A., Mensch, A., Bamford, C., Chaplot, D.~S., Casas, D. d.~l., Bressand, F., Lengyel, G., Lample, G., Saulnier, L., et~al.
\newblock Mistral 7b.
\newblock \emph{arXiv preprint arXiv:2310.06825}, 2023.

\bibitem[Kim et~al.(2023)Kim, Baldi, and McAleer]{kim2023language}
Kim, G., Baldi, P., and McAleer, S.
\newblock Language models can solve computer tasks.
\newblock \emph{arXiv preprint arXiv:2303.17491}, 2023.

\bibitem[LDJnr(2023)]{capybara_dataset}
LDJnr.
\newblock Capybara dataset.
\newblock \url{https://hf.co/datasets/LDJnr/Verified-Camel}, \url{https://hf.co/datasets/LDJnr/Pure-Dove}, \url{https://hf.co/datasets/LDJnr/LessWrong-Amplify-Instruct}, 2023.
\newblock A dataset focusing on reasoning in multi-turn conversations.

\bibitem[Lee et~al.(2022)Lee, Liang, and Yang]{lee2022coauthor}
Lee, M., Liang, P., and Yang, Q.
\newblock Coauthor: Designing a human-ai collaborative writing dataset for exploring language model capabilities.
\newblock In \emph{Proceedings of the 2022 CHI conference on human factors in computing systems}, pp.\  1--19, 2022.

\bibitem[Li et~al.(2023)Li, Song, Yu, Yu, Li, Huang, and Li]{li2023apibank}
Li, M., Song, F., Yu, B., Yu, H., Li, Z., Huang, F., and Li, Y.
\newblock Api-bank: A benchmark for tool-augmented llms, 2023.

\bibitem[Lian et~al.(2023)Lian, Goodson, Pentland, Cook, Vong, and "Teknium"]{OpenOrca}
Lian, W., Goodson, B., Pentland, E., Cook, A., Vong, C., and "Teknium".
\newblock Openorca: An open dataset of gpt augmented flan reasoning traces.
\newblock \url{https://https://huggingface.co/Open-Orca/OpenOrca}, 2023.

\bibitem[Liang et~al.(2022)Liang, Huang, Xia, Xu, Hausman, Ichter, Florence, and Zeng]{codeaspolicies2022}
Liang, J., Huang, W., Xia, F., Xu, P., Hausman, K., Ichter, B., Florence, P., and Zeng, A.
\newblock Code as policies: Language model programs for embodied control.
\newblock In \emph{arXiv preprint arXiv:2209.07753}, 2022.

\bibitem[Liu et~al.(2023{\natexlab{a}})Liu, Yuan, Fu, Jiang, Hayashi, and Neubig]{liu2023pre}
Liu, P., Yuan, W., Fu, J., Jiang, Z., Hayashi, H., and Neubig, G.
\newblock Pre-train, prompt, and predict: A systematic survey of prompting methods in natural language processing.
\newblock \emph{ACM Computing Surveys}, 55\penalty0 (9):\penalty0 1--35, 2023{\natexlab{a}}.

\bibitem[Liu et~al.(2023{\natexlab{b}})Liu, Yang, Jia, Zhang, Zhou, Dai, Yang, and Vosoughi]{liu2023training}
Liu, R., Yang, R., Jia, C., Zhang, G., Zhou, D., Dai, A.~M., Yang, D., and Vosoughi, S.
\newblock Training socially aligned language models in simulated human society.
\newblock \emph{arXiv preprint arXiv:2305.16960}, 2023{\natexlab{b}}.

\bibitem[Ma et~al.(2023)Ma, Liang, Wang, Huang, Bastani, Jayaraman, Zhu, Fan, and Anandkumar]{ma2023eureka}
Ma, Y.~J., Liang, W., Wang, G., Huang, D.-A., Bastani, O., Jayaraman, D., Zhu, Y., Fan, L., and Anandkumar, A.
\newblock Eureka: Human-level reward design via coding large language models.
\newblock \emph{arXiv preprint arXiv:2310.12931}, 2023.

\bibitem[Mialon et~al.(2023)Mialon, Dess{\`\i}, Lomeli, Nalmpantis, Pasunuru, Raileanu, Rozi{\`e}re, Schick, Dwivedi-Yu, Celikyilmaz, et~al.]{mialon2023augmented}
Mialon, G., Dess{\`\i}, R., Lomeli, M., Nalmpantis, C., Pasunuru, R., Raileanu, R., Rozi{\`e}re, B., Schick, T., Dwivedi-Yu, J., Celikyilmaz, A., et~al.
\newblock Augmented language models: a survey.
\newblock \emph{arXiv preprint arXiv:2302.07842}, 2023.

\bibitem[Nakano et~al.(2021)Nakano, Hilton, Balaji, Wu, Ouyang, Kim, Hesse, Jain, Kosaraju, Saunders, et~al.]{nakano2021webgpt}
Nakano, R., Hilton, J., Balaji, S., Wu, J., Ouyang, L., Kim, C., Hesse, C., Jain, S., Kosaraju, V., Saunders, W., et~al.
\newblock Webgpt: Browser-assisted question-answering with human feedback.
\newblock \emph{arXiv preprint arXiv:2112.09332}, 2021.

\bibitem[OpenChat(2023)]{sharegpt4_dataset}
OpenChat.
\newblock Sharegpt dataset.
\newblock \url{https://hf.co/datasets/openchat/openchat_sharegpt_v3/blob/main/sharegpt_gpt4.json}, 2023.
\newblock A dataset containing multi-turn conversations between human and LLM assistants. It is filtered to contain data only from GPT-4.

\bibitem[Ouyang et~al.(2022)Ouyang, Wu, Jiang, Almeida, Wainwright, Mishkin, Zhang, Agarwal, Slama, Ray, et~al.]{ouyang2022training}
Ouyang, L., Wu, J., Jiang, X., Almeida, D., Wainwright, C., Mishkin, P., Zhang, C., Agarwal, S., Slama, K., Ray, A., et~al.
\newblock Training language models to follow instructions with human feedback.
\newblock \emph{Advances in Neural Information Processing Systems}, 35:\penalty0 27730--27744, 2022.

\bibitem[Park et~al.(2023)Park, O'Brien, Cai, Morris, Liang, and Bernstein]{park2023generative}
Park, J.~S., O'Brien, J., Cai, C.~J., Morris, M.~R., Liang, P., and Bernstein, M.~S.
\newblock Generative agents: Interactive simulacra of human behavior.
\newblock In \emph{Proceedings of the 36th Annual ACM Symposium on User Interface Software and Technology}, pp.\  1--22, 2023.

\bibitem[Pasupat \& Liang(2015)Pasupat and Liang]{pasupat2015compositional}
Pasupat, P. and Liang, P.
\newblock Compositional semantic parsing on semi-structured tables.
\newblock In \emph{Proceedings of the 53rd Annual Meeting of the Association for Computational Linguistics and the 7th International Joint Conference on Natural Language Processing (Volume 1: Long Papers)}, pp.\  1470--1480, 2015.

\bibitem[Patil et~al.(2023)Patil, Zhang, Wang, and Gonzalez]{Patil2023GorillaLL}
Patil, S.~G., Zhang, T., Wang, X., and Gonzalez, J.~E.
\newblock Gorilla: Large language model connected with massive apis.
\newblock \emph{ArXiv}, abs/2305.15334, 2023.
\newblock URL \url{https://api.semanticscholar.org/CorpusID:258865184}.

\bibitem[Qian et~al.(2023)Qian, Cong, Yang, Chen, Su, Xu, Liu, and Sun]{qian2023communicative}
Qian, C., Cong, X., Yang, C., Chen, W., Su, Y., Xu, J., Liu, Z., and Sun, M.
\newblock Communicative agents for software development.
\newblock \emph{arXiv preprint arXiv:2307.07924}, 2023.

\bibitem[Qiao et~al.(2023)Qiao, Li, Zhang, He, Kang, Zhang, Yang, Dong, Zhang, Wang, et~al.]{qiao2023taskweaver}
Qiao, B., Li, L., Zhang, X., He, S., Kang, Y., Zhang, C., Yang, F., Dong, H., Zhang, J., Wang, L., et~al.
\newblock Taskweaver: A code-first agent framework.
\newblock \emph{arXiv preprint arXiv:2311.17541}, 2023.

\bibitem[Qin et~al.(2023{\natexlab{a}})Qin, Hu, Lin, Chen, Ding, Cui, Zeng, Huang, Xiao, Han, et~al.]{qin2023tool}
Qin, Y., Hu, S., Lin, Y., Chen, W., Ding, N., Cui, G., Zeng, Z., Huang, Y., Xiao, C., Han, C., et~al.
\newblock Tool learning with foundation models.
\newblock \emph{arXiv preprint arXiv:2304.08354}, 2023{\natexlab{a}}.

\bibitem[Qin et~al.(2023{\natexlab{b}})Qin, Liang, Ye, Zhu, Yan, Lu, Lin, Cong, Tang, Qian, Zhao, Tian, Xie, Zhou, Gerstein, Li, Liu, and Sun]{Qin2023ToolLLMFL}
Qin, Y., Liang, S., Ye, Y., Zhu, K., Yan, L., Lu, Y.-T., Lin, Y., Cong, X., Tang, X., Qian, B., Zhao, S., Tian, R., Xie, R., Zhou, J., Gerstein, M.~H., Li, D., Liu, Z., and Sun, M.
\newblock Toolllm: Facilitating large language models to master 16000+ real-world apis.
\newblock \emph{ArXiv}, abs/2307.16789, 2023{\natexlab{b}}.
\newblock URL \url{https://api.semanticscholar.org/CorpusID:260334759}.

\bibitem[Schick et~al.(2023)Schick, Dwivedi-Yu, Dess{\`\i}, Raileanu, Lomeli, Zettlemoyer, Cancedda, and Scialom]{schick2023toolformer}
Schick, T., Dwivedi-Yu, J., Dess{\`\i}, R., Raileanu, R., Lomeli, M., Zettlemoyer, L., Cancedda, N., and Scialom, T.
\newblock Toolformer: Language models can teach themselves to use tools.
\newblock \emph{arXiv preprint arXiv:2302.04761}, 2023.

\bibitem[Shen et~al.(2023)Shen, Song, Tan, Li, Lu, and Zhuang]{shen2023hugginggpt}
Shen, Y., Song, K., Tan, X., Li, D., Lu, W., and Zhuang, Y.
\newblock Hugginggpt: Solving ai tasks with chatgpt and its friends in huggingface.
\newblock \emph{arXiv preprint arXiv:2303.17580}, 2023.

\bibitem[Shridhar et~al.(2020)Shridhar, Yuan, Cote, Bisk, Trischler, and Hausknecht]{shridhar2020alfworld}
Shridhar, M., Yuan, X., Cote, M.-A., Bisk, Y., Trischler, A., and Hausknecht, M.
\newblock Alfworld: Aligning text and embodied environments for interactive learning.
\newblock In \emph{International Conference on Learning Representations}, 2020.

\bibitem[Singh et~al.(2023)Singh, Blukis, Mousavian, Goyal, Xu, Tremblay, Fox, Thomason, and Garg]{progprompt}
Singh, I., Blukis, V., Mousavian, A., Goyal, A., Xu, D., Tremblay, J., Fox, D., Thomason, J., and Garg, A.
\newblock Progprompt: Generating situated robot task plans using large language models.
\newblock In \emph{2023 IEEE International Conference on Robotics and Automation (ICRA)}, pp.\  11523--11530, 2023.
\newblock \doi{10.1109/ICRA48891.2023.10161317}.

\bibitem[Sur\'is et~al.(2023)Sur\'is, Menon, and Vondrick]{surismenon2023vipergpt}
Sur\'is, D., Menon, S., and Vondrick, C.
\newblock Vipergpt: Visual inference via python execution for reasoning.
\newblock \emph{Proceedings of IEEE International Conference on Computer Vision (ICCV)}, 2023.

\bibitem[Tang et~al.(2024)Tang, Jin, Zhu, Yuan, Zhang, Zhou, Qu, Zhao, Tang, Zhang, et~al.]{tang2024prioritizing}
Tang, X., Jin, Q., Zhu, K., Yuan, T., Zhang, Y., Zhou, W., Qu, M., Zhao, Y., Tang, J., Zhang, Z., et~al.
\newblock Prioritizing safeguarding over autonomy: Risks of llm agents for science.
\newblock \emph{arXiv preprint arXiv:2402.04247}, 2024.

\bibitem[{TIOBE Index}(2024)]{tiobe}
{TIOBE Index}.
\newblock Tiobe index.
\newblock \url{https://www.tiobe.com/tiobe-index/}, Accessed at Jan 23rd, 2024, 2024.
\newblock The TIOBE Programming Community index is an indicator of the popularity of programming languages. The index is updated once a month. The ratings are based on the number of skilled engineers world-wide, courses and third party vendors.

\bibitem[Touvron et~al.(2023)Touvron, Martin, Stone, Albert, Almahairi, Babaei, Bashlykov, Batra, Bhargava, Bhosale, et~al.]{touvron2023llama}
Touvron, H., Martin, L., Stone, K., Albert, P., Almahairi, A., Babaei, Y., Bashlykov, N., Batra, S., Bhargava, P., Bhosale, S., et~al.
\newblock Llama 2: Open foundation and fine-tuned chat models.
\newblock \emph{arXiv preprint arXiv:2307.09288}, 2023.

\bibitem[Wang et~al.(2023{\natexlab{a}})Wang, Xie, Jiang, Mandlekar, Xiao, Zhu, Fan, and Anandkumar]{wang2023voyager}
Wang, G., Xie, Y., Jiang, Y., Mandlekar, A., Xiao, C., Zhu, Y., Fan, L., and Anandkumar, A.
\newblock Voyager: An open-ended embodied agent with large language models.
\newblock \emph{arXiv preprint arXiv:2305.16291}, 2023{\natexlab{a}}.

\bibitem[Wang et~al.(2023{\natexlab{b}})Wang, Ma, Feng, Zhang, Yang, Zhang, Chen, Tang, Chen, Lin, et~al.]{wang2023survey}
Wang, L., Ma, C., Feng, X., Zhang, Z., Yang, H., Zhang, J., Chen, Z., Tang, J., Chen, X., Lin, Y., et~al.
\newblock A survey on large language model based autonomous agents.
\newblock \emph{arXiv preprint arXiv:2308.11432}, 2023{\natexlab{b}}.

\bibitem[Wang et~al.(2022{\natexlab{a}})Wang, Jansen, C{\^o}t{\'e}, and Ammanabrolu]{Wang2022ScienceWorldIY}
Wang, R., Jansen, P.~A., C{\^o}t{\'e}, M.-A., and Ammanabrolu, P.
\newblock Scienceworld: Is your agent smarter than a 5th grader?
\newblock In \emph{Conference on Empirical Methods in Natural Language Processing}, 2022{\natexlab{a}}.
\newblock URL \url{https://api.semanticscholar.org/CorpusID:247451124}.

\bibitem[Wang et~al.(2022{\natexlab{b}})Wang, Wei, Schuurmans, Le, Chi, Narang, Chowdhery, and Zhou]{wang2022self}
Wang, X., Wei, J., Schuurmans, D., Le, Q., Chi, E., Narang, S., Chowdhery, A., and Zhou, D.
\newblock Self-consistency improves chain of thought reasoning in language models.
\newblock \emph{arXiv preprint arXiv:2203.11171}, 2022{\natexlab{b}}.

\bibitem[Wang et~al.(2023{\natexlab{c}})Wang, Li, and Ji]{wang-etal-2023-code4struct}
Wang, X., Li, S., and Ji, H.
\newblock {C}ode4{S}truct: Code generation for few-shot event structure prediction.
\newblock In Rogers, A., Boyd-Graber, J., and Okazaki, N. (eds.), \emph{Proceedings of the 61st Annual Meeting of the Association for Computational Linguistics (Volume 1: Long Papers)}, pp.\  3640--3663, Toronto, Canada, July 2023{\natexlab{c}}. Association for Computational Linguistics.
\newblock \doi{10.18653/v1/2023.acl-long.202}.
\newblock URL \url{https://aclanthology.org/2023.acl-long.202}.

\bibitem[Wang et~al.(2023{\natexlab{d}})Wang, Peng, Jabbarvand, and Ji]{Wang2023LeTI}
Wang, X., Peng, H., Jabbarvand, R., and Ji, H.
\newblock Leti: Learning to generate from textual interactions.
\newblock \emph{ArXiv}, abs/2305.10314, 2023{\natexlab{d}}.

\bibitem[Wang et~al.(2023{\natexlab{e}})Wang, Wang, Liu, Chen, Yuan, Peng, and Ji]{wang2023mint}
Wang, X., Wang, Z., Liu, J., Chen, Y., Yuan, L., Peng, H., and Ji, H.
\newblock Mint: Evaluating llms in multi-turn interaction with tools and language feedback.
\newblock \emph{arXiv preprint arXiv:2309.10691}, 2023{\natexlab{e}}.

\bibitem[Wang et~al.(2023{\natexlab{f}})Wang, Cai, Liu, Ma, and Liang]{wang2023describe}
Wang, Z., Cai, S., Liu, A., Ma, X., and Liang, Y.
\newblock Describe, explain, plan and select: Interactive planning with large language models enables open-world multi-task agents.
\newblock \emph{arXiv preprint arXiv:2302.01560}, 2023{\natexlab{f}}.

\bibitem[Wei et~al.(2022)Wei, Wang, Schuurmans, Bosma, Xia, Chi, Le, Zhou, et~al.]{wei2022chain}
Wei, J., Wang, X., Schuurmans, D., Bosma, M., Xia, F., Chi, E., Le, Q.~V., Zhou, D., et~al.
\newblock Chain-of-thought prompting elicits reasoning in large language models.
\newblock \emph{Advances in Neural Information Processing Systems}, 35:\penalty0 24824--24837, 2022.

\bibitem[Xu et~al.(2023)Xu, Hong, Li, Hu, Chen, and Zhang]{xu2023tool}
Xu, Q., Hong, F., Li, B., Hu, C., Chen, Z., and Zhang, J.
\newblock On the tool manipulation capability of open-source large language models, 2023.

\bibitem[Yang et~al.(2024{\natexlab{a}})Yang, Prabhakar, Narasimhan, and Yao]{yang2024intercode}
Yang, J., Prabhakar, A., Narasimhan, K., and Yao, S.
\newblock Intercode: Standardizing and benchmarking interactive coding with execution feedback.
\newblock \emph{Advances in Neural Information Processing Systems}, 36, 2024{\natexlab{a}}.

\bibitem[Yang et~al.(2024{\natexlab{b}})Yang, Liu, Wu, Yang, Fung, Li, Huang, Cao, Wang, Wang, Ji, and Zhai]{yang2024llm}
Yang, K., Liu, J., Wu, J., Yang, C., Fung, Y.~R., Li, S., Huang, Z., Cao, X., Wang, X., Wang, Y., Ji, H., and Zhai, C.
\newblock If llm is the wizard, then code is the wand: A survey on how code empowers large language models to serve as intelligent agents, 2024{\natexlab{b}}.

\bibitem[Yang et~al.(2018)Yang, Qi, Zhang, Bengio, Cohen, Salakhutdinov, and Manning]{yang2018hotpotqa}
Yang, Z., Qi, P., Zhang, S., Bengio, Y., Cohen, W., Salakhutdinov, R., and Manning, C.~D.
\newblock Hotpotqa: A dataset for diverse, explainable multi-hop question answering.
\newblock In \emph{Proceedings of the 2018 Conference on Empirical Methods in Natural Language Processing}, pp.\  2369--2380, 2018.

\bibitem[Yang et~al.(2024{\natexlab{c}})Yang, Liu, Liu, Liu, Xiong, Wang, Yang, Hu, Chen, Zhang, Luo, Guo, Li, and Liu]{yang2024unified}
Yang, Z., Liu, A., Liu, Z., Liu, K., Xiong, F., Wang, Y., Yang, Z., Hu, Q., Chen, X., Zhang, Z., Luo, F., Guo, Z., Li, P., and Liu, Y.
\newblock Towards unified alignment between agents, humans, and environment, 2024{\natexlab{c}}.

\bibitem[Yao et~al.(2022{\natexlab{a}})Yao, Chen, Yang, and Narasimhan]{yao2022webshop}
Yao, S., Chen, H., Yang, J., and Narasimhan, K.
\newblock Webshop: Towards scalable real-world web interaction with grounded language agents.
\newblock \emph{Advances in Neural Information Processing Systems}, 35:\penalty0 20744--20757, 2022{\natexlab{a}}.

\bibitem[Yao et~al.(2022{\natexlab{b}})Yao, Zhao, Yu, Du, Shafran, Narasimhan, and Cao]{yao2022react}
Yao, S., Zhao, J., Yu, D., Du, N., Shafran, I., Narasimhan, K.~R., and Cao, Y.
\newblock React: Synergizing reasoning and acting in language models.
\newblock In \emph{The Eleventh International Conference on Learning Representations}, 2022{\natexlab{b}}.

\bibitem[Yao et~al.(2023{\natexlab{a}})Yao, Yu, Zhao, Shafran, Griffiths, Cao, and Narasimhan]{yao2023tree}
Yao, S., Yu, D., Zhao, J., Shafran, I., Griffiths, T.~L., Cao, Y., and Narasimhan, K.
\newblock Tree of thoughts: Deliberate problem solving with large language models.
\newblock \emph{arXiv preprint arXiv:2305.10601}, 2023{\natexlab{a}}.

\bibitem[Yao et~al.(2023{\natexlab{b}})Yao, Heinecke, Niebles, Liu, Feng, Xue, Murthy, Chen, Zhang, Arpit, et~al.]{yao2023retroformer}
Yao, W., Heinecke, S., Niebles, J.~C., Liu, Z., Feng, Y., Xue, L., Murthy, R., Chen, Z., Zhang, J., Arpit, D., et~al.
\newblock Retroformer: Retrospective large language agents with policy gradient optimization.
\newblock \emph{arXiv preprint arXiv:2308.02151}, 2023{\natexlab{b}}.

\bibitem[Yuan et~al.(2023)Yuan, Chen, Wang, Fung, Peng, and Ji]{Yuan2023CRAFTCL}
Yuan, L., Chen, Y., Wang, X., Fung, Y.~R., Peng, H., and Ji, H.
\newblock Craft: Customizing llms by creating and retrieving from specialized toolsets.
\newblock \emph{ArXiv}, abs/2309.17428, 2023.
\newblock URL \url{https://api.semanticscholar.org/CorpusID:263310662}.

\bibitem[Zeng et~al.(2023)Zeng, Liu, Lu, Wang, Liu, Dong, and Tang]{zeng2023agenttuning}
Zeng, A., Liu, M., Lu, R., Wang, B., Liu, X., Dong, Y., and Tang, J.
\newblock Agenttuning: Enabling generalized agent abilities for llms, 2023.

\bibitem[Zhang et~al.(2023)Zhang, Liu, Wang, Wang, Sun, Wang, and Cai]{zhang2023prefer}
Zhang, C., Liu, L., Wang, J., Wang, C., Sun, X., Wang, H., and Cai, M.
\newblock Prefer: Prompt ensemble learning via feedback-reflect-refine.
\newblock \emph{arXiv preprint arXiv:2308.12033}, 2023.

\bibitem[Zheng et~al.(2023)Zheng, Chiang, Sheng, Zhuang, Wu, Zhuang, Lin, Li, Li, Xing, et~al.]{zheng2023judging}
Zheng, L., Chiang, W.-L., Sheng, Y., Zhuang, S., Wu, Z., Zhuang, Y., Lin, Z., Li, Z., Li, D., Xing, E., et~al.
\newblock Judging llm-as-a-judge with mt-bench and chatbot arena.
\newblock \emph{arXiv preprint arXiv:2306.05685}, 2023.

\bibitem[Zheng et~al.(2024)Zheng, Zhang, Shen, Liu, Lin, Fu, Chen, and Yue]{opencodeinterpreter}
Zheng, T., Zhang, G., Shen, T., Liu, X., Lin, B.~Y., Fu, J., Chen, W., and Yue, X.
\newblock Opencodeinterpreter: Integrating code generation with execution and refinement.
\newblock \emph{https://arxiv.org/abs/2402.14658}, 2024.

\bibitem[Zhu et~al.(2023)Zhu, Chen, Tian, Tao, Su, Yang, Huang, Li, Lu, Wang, et~al.]{zhu2023ghost}
Zhu, X., Chen, Y., Tian, H., Tao, C., Su, W., Yang, C., Huang, G., Li, B., Lu, L., Wang, X., et~al.
\newblock Ghost in the minecraft: Generally capable agents for open-world enviroments via large language models with text-based knowledge and memory.
\newblock \emph{arXiv preprint arXiv:2305.17144}, 2023.

\end{thebibliography}
\bibliographystyle{icml2024}

\appendix
\newpage
\renewcommand{\thefigure}{A.\arabic{figure}}
\renewcommand{\thetable}{A.\arabic{table}}
\newcounter{tcolorboxcounter}
\renewcommand{\thetcolorboxcounter}{\arabic{tcolorboxcounter}}
\renewcommand{\thetcolorboxcounter}{A.\arabic{tcolorboxcounter}}  
\onecolumn

\begin{table}[h]
    \centering
    \caption{Example of actions for re-purposed API-Bank \citep{li2023apibank} and \evalname.}
    \resizebox{0.7\columnwidth}{!}{%
    \begin{tabular}{@{} l p{8cm} @{}}
        \toprule
        \textbf{Format} & \textbf{Action} \\
        \midrule
        \multirow{2}*{\approach} & \texttt{AddAgenda(content="Meeting with John", time="2023-10-26 09:00:00")} \\
        \hline
        \multirow{3}*{JSON} & \texttt{ \{"action": "AddAgenda", "content": "Meeting with John", "time": "2023-10-26 09:00:00"\} } \\ 
        \hline
        \multirow{2}*{Text} & \texttt{Action: AddAgenda, content: Meeting with John, time: 2023-10-26 09:00:00} \\
        \bottomrule
        \end{tabular}
    }
    \label{tab:apibank-action-example}
    \vspace{-0.3cm}
\end{table}

\begin{table*}[!htb]
\centering
\caption{Comparison between \evalname and existing tool-use evaluation benchmark.
}

\resizebox{\textwidth}{!}{
\begin{threeparttable}
\begin{tabular}{l|ccccc}
\toprule
\textbf{Benchmark} & \evalname & ToolBench & APIBench & API-Bank & ToolBench \\
& (This work) & \citep{Qin2023ToolLLMFL} & \citep{Patil2023GorillaLL} & \citep{li2023apibank} & \citep{xu2023tool} \\
\midrule
Requiring multi-turn interaction       & \greencheck & \greencheck & \redwrong & \redwrong & \redwrong \\
Multiple tools                         & \greencheck & \greencheck & \greencheck & \greencheck & \greencheck \\
Evaluation                     & Answer Match & LLM Evaluator & AST Tree Match & API-Call Match & Test Case \\
No dependency on external API$^*$        & \greencheck & \redwrong & \redwrong & \greencheck & \redwrong \\
Supported API Action Format             & \approach \& JSON \& Text & JSON & \approach & JSON & \approach \\
\bottomrule
\end{tabular}
\begin{tablenotes}
    \item[*] Whether to rely on external API (e.g., RapidAPI, Google Sheet) hosted by a third party. The availability of such third-party APIs can greatly impact evaluation results (e.g., low API-calling performance not because the model is bad but rather because the API required is not accessible).
\end{tablenotes}
\end{threeparttable}
}
\label{tab:tool_bench_comparison}
\end{table*}

\begin{table*}[!h]
    \centering
\caption{Ablation study results. The best results are \textbf{bolded}, and the second-best results are \underline{underlined}.
ID and OD stand for in-domain and out-of-domain evaluation correspondingly.
Overall averaged performance normalizes the MT-Bench score to be consistent with other tasks and excludes in-domain tasks for fair comparison.
}
\label{tab:ablation_study_results}

\begin{adjustbox}{max width=\textwidth}
\begin{tabular}{@{} lc rr m{0.01em} rr m{0.01em} rrrr m{0.01em} r@{}}
\toprule
& & \multicolumn{5}{c}{\textbf{Agent Tasks}}                                                        && \multicolumn{4}{c}{\textbf{Generic LLM Tasks}} && \textbf{Overall} \\
 \cmidrule{3-7}
& & \multicolumn{2}{c}{\textbf{Code as Action}} && \multicolumn{2}{c}{\textbf{Text as Action (OD)}} && \multicolumn{4}{c}{\textbf{(OD)}} && \textbf{Average} \\
  \cmidrule{3-4}
  \cmidrule{6-7}
  \cmidrule{9-12}
  Model & Size & MINT (ID) & MINT (OD) && Miniwob++ & SciWorld && MMLU & HumanEval & GSM8K & MTBench \\
\midrule

\modelname (Llama2-based) & 7B & $\mathbf{51.3}$ & $\mathbf{20.4}$ &  & \underline{$25.5$} & $\mathbf{17.6}$ &  & $\mathbf{50.6}$ & \underline{$18.1$} & $\mathbf{38.3}$ & $\mathbf{7.5}$ &  & $\mathbf{35.1}$ \\
w/o \approach & 7B & $17.0$ & $15.5$ &  & $\mathbf{36.4}$ & $16.9$ &  & \underline{$49.5$} & $14.7$ & \underline{$36.0$} & \underline{$7.2$} &  & \underline{$34.5$} \\
w/o general conversations & 7B & \underline{$29.2$} & \underline{$15.9$} &  & $0.0$ & \underline{$17.1$} &  & $46.4$ & $\mathbf{19.7}$ & $20.6$ & $4.1$ &  & $22.9$ \\

\midrule

\modelname (Mistral-based) & 7B & $\mathbf{57.4}$ & $\mathbf{32.4}$ &  & \underline{$46.2$} & \underline{$15.9$} &  & \underline{$59.1$} & $\mathbf{34.7}$ & \underline{$58.0$} & \underline{$8.2$} &  & $\mathbf{46.8}$ \\
w/o \approach & 7B & $32.9$ & \underline{$23.0$} &  & $\mathbf{47.8}$ & $\mathbf{17.0}$ &  & $\mathbf{59.9}$ & \underline{$33.2$} & $\mathbf{59.5}$ & $\mathbf{8.3}$ &  & \underline{$46.2$} \\
w/o general conversations & 7B & \underline{$50.5$} & $13.9$ &  & $0.0$ & $11.0$ &  & $52.4$ & $27.9$ & $26.8$ & $2.6$ &  & $22.6$ \\
\bottomrule
\end{tabular}
\end{adjustbox}
\end{table*}

\section{Comparison with Work that Uses Code Generation for Problem-solving}
\label{sec:prior_code_action_work}

In this section, we discuss the fundamental differences between \approach and prior work that prompt LLM to generate code for problem-solving. 
Existing work have explored using code generation for task-solving in different domains, for example, Code4Struct \citep{wang-etal-2023-code4struct} for structured prediction, PaL \citep{gao2023pal} for math reasoning, Meta-GPT \citep{hong2023metagpt} for multi-agent collaboration, code-as-policy \citep{codeaspolicies2022} for robot control, ViperGPT \citep{surismenon2023vipergpt} for visual question answering, Voyager \citep{wang2023voyager} for playing games, Data Interpreter \citep{hong2024data} for data science tasks, etc.

Most prior work generates code (i.e., a static sequence of actions) in a single-turn setting and cannot dynamically readjust action on new observation: It is considered a failure when the model-generated code fails to solve a task on the \textit{first attempt}.
This setting overlooks the potential of environmental observation (e.g., code execution results) that might benefit future action and overall decision (e.g., dynamically adjusting subsequent code after observing intermediate code execution results, fixing erroneous code after seeing an error message).
That is, the generated code is a static sequence of actions that cannot be dynamically re-adjusted on the fly by incorporating new observations.
Such a single-turn setting makes it challenging to scale to more challenging problems since even expert human programmers usually cannot write functionally correct code in the first pass.
On the other hand, \approach is a multi-turn interaction agent framework that allows dynamic adjustment of prior actions or emitting new actions by design (\sref{sec:codeact_definition}, \fref{fig:llm_agent_framework}) and is compatible with any form of textual observation (e.g., tool execution output, automated feedback) from the environment.
Beyond being compatible with environmental observation, our instruction tuning dataset \dataname specifically collects data for multi-turn self-improving, offering a practical solution to enhance LLM's multi-turn self-improving process.

In addition, previous approaches require heavy prompt engineering and crafting of few-shot demonstrations to tailor LLMs to a particular domain or task (e.g., robot control~\citep{codeaspolicies2022}) since the backbone LLMs are not specially optimized for dynamic planning and decision making.
In contrast, in this work, we propose the \approach framework that uses executable Python code to consolidate LLM agents' actions into unified action space and collect \dataname on a diverse array of tasks (e.g., information seeking, tabular reasoning, robot planning, etc) to make the trained model, \modelname, easily scale to diverse tasks and domains with minimal human efforts as shown in \sref{sec:llm_agent_evaluation}. 

One notable exception among prior work is Voyager \citep{wang2023voyager}, which performs iterative prompting in a constrained action space of \textit{function definitions} to fix code errors.
Different from \approach, such setting disallows dynamic re-adjustment of \textit{atomic} actions on the fly: In \approach, for a particular task (e.g., craft stone sword in Minecraft), the agent can first execute one line of code (any atomic action or composed functions, e.g., move forward, locate stone), and dynamically produce different actions based on the observation of the first action.
This is challenging for Voyager to achieve: Similar to code-as-policy \citep{codeaspolicies2022}, they generate action (a skill, e.g., craft stone sword) as a Python \textit{function definition} that outlines the entire plan for a task (e.g., multi-step code outlining how you should craft a stone sword and handles for different potential cases, which requires strong domain knowledge).
This imposes significant constraints on the agent's action space and disallows dynamic re-adjustment of \textit{atomic} actions on the fly: That is, the agent can only generate one complete function first (e.g., by imaging all possible cases that might happen when you try to locate stones), execute the entire function, observe the feedback, and update the entire function as action in the subsequent move.
Besides the constrained ability to re-adjust action from environmental observation, they also rely on heavy prompting engineering (a typical drawback discussed above) to provide relevant information (e.g., current state, additional self-critics via prompting) to generate revised code, whereas \approach is situated in a setting that requires no prompt engineering efforts: the context window of LLM only contains its \textit{past actions and observations} and does not require human efforts to filter for relevant information.

Similar to \approach, concurrent work OpenCodeInterpreter \citep{opencodeinterpreter}, with a specific focus on competitive code generation questions, collects code-debugging trajectories to improve an LLM's iterative code debugging performance. However, its applicability to general LLM agent tasks remains unknown.

\section{Comparison with TaskWeaver}
\label{sec:taskweaver}
In the landscape of unifying the action space of LLM agents, our work represents a leap over the previous initiative, TaskWeaver~\citep{qiao2023taskweaver}. 
While TaskWeaver deserves acknowledgment for initially integrating code into the action space of LLM agents, its exploration remains limited.
This work, primarily characterized by its reliance on a limited set of qualitative examples with close-sourced models as the backbones, fails to harness the full potential of this integration, remaining merely conceptual demonstrations.
Our work transcends mere conceptualization by conducting an extensive and rigorous analysis, clearly quantifying the benefits of code action within LLM agents.
Beyond this, we introduce a unique instruction-tuning dataset \dataname specifically designed to amplify the agent's capabilities in executing code-based actions and an open-source LLM agent \modelname.
These contributions not only extend the work of TaskWeaver but also pave the way for future explorations, offering valuable resources to the open-source community and redefining the potential of LLM agents in practical applications.

\section{General Data Down-sample}
\label{sec:general_data_downsample}

\begin{itemize}
    \item \textbf{ShareGPT} \citep{sharegpt_dataset}: We remove all single-turn conversations, then perform random sub-sample to a desired final number.
    \item \textbf{ShareGPT (GPT-4)} \citep{sharegpt4_dataset}: We do not perform sub-sampling on this dataset.
    \item \textbf{OpenOrca} \citep{OpenOrca}: We select the CoT subset of OpenOrca, then perform a random sub-sample to a desired final number.
    \item \textbf{CapyBara} \citep{capybara_dataset}: We do not perform sub-sampling on this dataset.
\end{itemize}

\section{\modelname Training Details}
\label{sec:model_training_details}

All SFT experiments are performed on one 4xA100 40GB SXM node using a fork of Megatron-LLM \citep{epfmgtrn} with a training throughput of around 9k tokens per second. We use chatML format\footnote{\url{https://github.com/openai/openai-python/blob/release-v0.28.0/chatml.md}} for all multi-turn data, and we only calculate and optimize for loss on the assistant response. We pack short instances into longer ones and apply flash attention for training efficiency. 

We train both LLaMA-2 and Mistral LLMs with Tensor Parallel of 4, the learning rate of \texttt{1e-5} with 50 warmup steps and cosine decay (end learning rate of \texttt{1e-6}). We train for five epochs with a batch size of 32. We use the 3rd epoch checkpoint for all our experiments.

\section{Example Prompt for \approach}
\label{sec:codeact_prompt}

This is an example (zero-shot) system prompt used in a deploy instance of \approach where we used chatML format.

The users may optionally include tools descriptions similar to \sref{sec:zeroshot_act_prompt} or including extra in-context examples similar to \sref{sec:interaction_data_gen_prompt}.

\begin{lstlisting}[breaklines]
<|im_start|>system
A chat between a curious user and an artificial intelligence assistant. The assistant gives helpful, detailed, and polite answers to the user's questions.
The assistant can interact with an interactive Python (Jupyter Notebook) environment and receive the corresponding output when needed. The code should be enclosed using "<execute>" tag, for example: <execute> print("Hello World!") </execute>.
The assistant should attempt fewer things at a time instead of putting too much code in one <execute> block. The assistant can install packages through PIP by <execute> !pip install [package needed] </execute> and should always import packages and define variables before starting to use them.
The assistant should stop <execute> and provide an answer when they have already obtained the answer from the execution result. Whenever possible, execute the code for the user using <execute> instead of providing it.
The assistant's response should be concise, but do express their thoughts.
<|im_end|>
\end{lstlisting}

\section{\evalname Prompt}
\label{sec:zeroshot_act_prompt}

\begin{lstlisting}[breaklines]
You have access to the following tools:
{{Tool Definition}}

{{Formatting Instruction}}

Now, let's get started!

Instruction: {{Example: Find the current price of Legendary Wand.}}
Answer in the format of 'xx.xx' (e.g., 12.34).

You can optionally express your thoughts using natural language before your action. For example, 'Thought: I want to use tool_name to do something. Action: <your action to call tool_name> End Action'.
Note that your output should always contain either 'Action:' or 'Answer:', but not both.
When you are done, output the result using 'Answer: your answer'
Please ONLY output the answer (e.g., single number), without any other text.
\end{lstlisting}

Each \texttt{\{\{...\}\}} component above will be substituted with corresponding information.

\subsection{Example of \{\{Tool Definition\}\}}

The following is an example tool definition for web-browsing.

\begin{lstlisting}[breaklines]
[1] click_url: Clicks on a URL. A clickable URL looks like [Clickable '<url_argument>'] in the webpage.
Arguments: url (str).
Returns the rendered content of the webpage after clicking the URL showing on the current rendered page.
    Signature: click_url(url: str) -> str
[2] go_to_previous_page: Goes back to the previous page. It has no arguments.
After going back to the previous page, return the rendered content of the webpage.
    Signature: go_to_previous_page() -> str
[3] scroll_down: Scrolls down the view. It has no arguments.
Returns the rendered content of the webpage after scrolling down.
    Signature: scroll_down() -> str
[4] scroll_up: Scrolls up the view. It has no arguments.
Returns the rendered content of the webpage after scrolling up.
    Signature: scroll_up() -> str
[5] view: Return the current view in string format of the rendered webpage. It has no arguments.
Returns the rendered content of the webpage.
You should call this when you want to see the rendered content of the current webpage.
    Signature: view() -> str
[6] calculator: Evaluates the given expression and returns the result. Accepts a calculation expression as input. For example, "2 + (3 * 4)" will return 14.
    Signature: calculator(expression: str) -> float
\end{lstlisting}

\subsection{Example of \{\{Formatting Instruction\}\}}

Different action format has different formatting instructions.

\subsection{Formatting Instruction for Code as Action}

\begin{lstlisting}[breaklines]
You can use the tools by outputing a block of Python code that invoke the tools.
You may use for-loops, if-statements, and other Python constructs when necessary.
Be sure to print the final answer at the end of your code.
You should begin your tool invocation with 'Action:' and end it with 'End Action'.
Example: 'Action:
tool_name(argument_1)
End Action'
\end{lstlisting}

\subsection{Formatting Instruction for Json as Action}

\begin{lstlisting}[breaklines]
You can use the tools by outputing a JSON object with the following fields:
  - 'tool': the name of the tool
  - 'args': a list of arguments to the tool
You should begin your tool invocation with 'Action:' and end it with 'End Action'.
Example: 'Action: {"tool": "tool_name", "args": ["argument_1"]} End Action'
You can only invoke one tool at a time.
\end{lstlisting}

\subsection{Formatting Instruction for Text as Action}

\begin{lstlisting}[breaklines]
You can use the tools by outputing the tool name followed by its arguments, delimited by commas.
You should begin your tool invocation with 'Action:' and end it with 'End Action'.
Example: 'Action: tool_name, argument_1 End Action'
You can only invoke one tool at a time.
\end{lstlisting}

\section{\approach Interaction Data}

\subsection{Dataset Downsample}
\label{sec:how_to_downsample_dataset}

\begin{table*}[!htb]
\centering
\caption{\dataname components and the number of instances for training trajectory generation.}
\resizebox{\textwidth}{!}{
\begin{tabular}{ll|lr}
\toprule
Domain & Capability & Dataset & \# of Instances \\
\midrule
Web Search & Information seeking through search API & HotpotQA \citep{yang2018hotpotqa} & 3,000 \\
Math Reasoning & Math problem-solving using math Libraries in Python (e.g., \texttt{sympy}) & MATH \citep{hendrycks2021measuring} & 5,586 \\
Code Generation & Self-debug from Python error messages and traceback & APPS \citep{hendrycks2021math} & 4,439 \\
Tabular Reasoning & Tabular Reasoning using \texttt{pandas} and \texttt{sqlite3} (for SQL) library & WikiTableQuestion \citep{pasupat2015compositional} & 3,000 \\
Embodied Planning & Interact with embodied environments through APIs & ALFWorld \citep{shridhar2020alfworld} & 3,553 \\
\bottomrule
\end{tabular}
}
\label{tab:data_training_instances}
\end{table*}

\begin{itemize}
    \item \textbf{Code generation tasks in APPS} \citep{hendrycks2021measuring}: We remove instances without any test case available.

    \item \textbf{Tabular reasoning tasks in WikiTableQuestion} \citep{pasupat2015compositional}: We select a subset of 3000 instances with the largest table size (i.e., sort by number of rows and columns) from the original dataset (14149 instances), and randomly assign 1500 of them to be \texttt{pandas}-based problems, and the rest 1500 to be SQL-based problems.

    \item \textbf{Web search tasks in HotpotQA} \citep{yang2018hotpotqa}: We select the 15661 problems labeled as ``hard" in the original dataset (with 90447 instances), then randomly down-sample them to 3000 problems.

    \item \textbf{Math reasoning in MATH} \citep{hendrycks2021math}: We remove problems with the annotated difficulty lower than 3, which results in 5586 instances as shown in \tref{tab:data_training_instances}.

    \item \textbf{Embodied Planning in ALFWorld} \citep{shridhar2020alfworld}: We did not perform down-sampling for AlfWorld.
    
\end{itemize}

\subsection{Data Selection Heuristic}
\label{sec:data_filter_heuristic}

Given successful task-solving trajectories that have more than 2 turns, we apply the following heuristic to select instances that can promote the code-as-actions, self-improvement, and instruction-following capabilities of LLM agents: 

\begin{itemize}
    \item \textbf{Code-as-Actions:} 
    We exclude trajectories wherein LLM agents do not adhere to the code-as-actions framework, either due to incorrect API invocation or the generation of actions in formats unsuitable for parsing and execution.

    \item \textbf{Self-Improving:} We selectively preserve those trajectories wherein the model initially encounters errors but subsequently rectifies these inaccuracies in later interactions. In addition, we eliminate successful trajectories that exclusively yield errors in all code executions. These are deemed ineffective demonstrations, as our objective is to prevent the model from learning to consistently execute erroneous code while still managing to provide correct answers.
    
    \item \textbf{Instruction-Following: }We remove rare cases where the LLM agents fail to follow the instruction and respond to the user, identified by an odd number of interaction turns.

\end{itemize}

After applying all these heuristics, we obtain 6728 trajectories (out of 6985) from \texttt{gpt-3.5} and \texttt{claude}, and 411 trajectories (out of 413) from \texttt{gpt-4-0613}.

\subsection{Example of Trajectory Generation Prompt}
\label{sec:interaction_data_gen_prompt}

The format of the data generation prompt closely follow MINT \citep{wang2023mint}.

\subsubsection{Tabular Reasoning (WikiTableQuestion)}
\label{sec:interaction_data_gen_prompt_tabular}
We only provide one-shot example for SQL-based tabular reasoning. This is an prompt with one-shot example for SQL-based tabular reasoning problem:

\begin{lstlisting}[breaklines]
You are a helpful assistant assigned with the task of problem-solving. To achieve this, you will be using an interactive coding environment equipped with a variety of tool functions to assist you throughout the process.

At each turn, you should first provide your step-by-step thinking for solving the task. Your thought process should be enclosed using "<thought>" tag, for example: <thought> I need to print "Hello World!" </thought>.

After that, you have two options:

1) Interact with a Python programming environment and receive the corresponding output. Your code should be enclosed using "<execute>" tag, for example: <execute> print("Hello World!") </execute>.
2) Directly provide a solution that adheres to the required format for the given task. Your solution should be enclosed using "<solution>" tag, for example: The answer is <solution> A </solution>.

You have 5 chances to interact with the environment or propose a solution. You can only propose a solution 2 times.



---

Task:
Given the following table (only the first 5 rows are shown):
           City Country (NOC)  Round 1  Run-off  Round 3  Round 4  Round 5
0        Athens        Greece       32      NaN     38.0     52.0     66.0
1          Rome         Italy       23      NaN     28.0     35.0     41.0
2     Cape Town  South Africa       16     62.0     22.0     20.0      NaN
3     Stockholm        Sweden       20      NaN     19.0      NaN      NaN
4  Buenos Aires     Argentina       16     44.0      NaN      NaN      NaN

Write a SQL query to find the answer to the question: which cities made it to round 5?.
The SQLite3 database is preloaded for you and can be accessed within <execute> block via the variable `conn` (SQLite3 connection object).
The table name is `data_table`.

Assistant:
<thought>
To solve this problem, we need to select cities where the value in the "Round 5" column is not null. Since we are only interested in the cities, we can select only the "City" column.
</thought>
<execute>
query = """
SELECT 
    City
FROM 
    data_table
WHERE 
    `Round 5` IS NOT NULL;
"""
# conn is preloaded
result = conn.execute(query).fetchall()
print(result)
</execute>

Observation:
[('Athens',), ('Rome',)]

Assistant:
The answer is <solution> ["Athens", "Rome"] </solution>.


---

Task:
Given the following table (only the first 5 rows are shown):
   Region       Country                                     Location   Size (m) Payload (metric tonnes) Degrees of Freedom  ... Z Vert vel (mm/s) X Horiz accel (m/s2) Y Horiz accel (m/s2) Z Vert accel (m/s2) Max Freq (Hz) Details checked
   +-
0  Africa       Algeria             CGS Laboratory (in construction)  6.1 x 6.1                      60                  6  ...             +-1000                  +-10                  +-10                  +-8           100       30/6/2010
1  Africa  South Africa                  University of Witwatersrand      4 x 4                      10                  1  ...               NaN                  +-10                  NaN                 NaN            40       17/7/2009
2    Asia         China  China Academy of Building Research, Beijing  6.1 x 6.1                      60                  6  ...              +-800                  +-15                  +-10                  +-8            50               ?
3    Asia         China                         Guangzhou University      3 x 3                      20                  6  ...             +-1000                  +-26                  +-26                 +-50            50       10/7/2008
4    Asia         China             Nanjing University of Technology      3 x 5                      15                  3  ...              +-500                  +-10                  +-10                 +-10            50               ?

[5 rows x 17 columns]

Write a SQL query to find the answer to the question: which is the other besides asia the most region charted.
The SQLite3 database is preloaded for you and can be accessed within <execute> block via the variable `conn` (SQLite3 connection object).
\end{lstlisting}

This is an example instruction for Pandas-package-based\footnote{\url{https://pandas.pydata.org/}} tabular reasoning problem:

\begin{lstlisting}[breaklines]
Task:
Given the following table (only the first 5 rows are shown):
  Pos  No           Rider     Bike  Laps       Time  Grid  Points
0   1  93    Marc Marquez    Derbi  22.0  40:46.315     1    25.0
1   2  38   Bradley Smith  Aprilia  22.0     +4.638     3    20.0
2   3  44   Pol Espargaro    Derbi  22.0     +4.996     2    16.0
3   4  11  Sandro Cortese    Derbi  22.0    +45.366     5    13.0
4   5   7   Efren Vazquez    Derbi  22.0    +45.433     8    11.0

Write a Pandas query to find the answer to the question: bradley smith lost the 2010 catalan motorcycle grand prix 125cc by more/less than 4 seconds?.
The dataframe is preloaded for you and can be accessed within <execute> block via the variable `df`.    
\end{lstlisting}

\subsubsection{Code Generation (APPS)}
\label{sec:interaction_data_gen_prompt_code}

Here is an example of the prompt with one in-context example for code generation on the APPS dataset \citep{hendrycks2021measuring} that encourages the LLM to self-debug its solution:

\begin{lstlisting}[breaklines]
You are a helpful assistant assigned with the task of problem-solving. To achieve this, you will be using an interactive coding environment equipped with a variety of tool functions to assist you throughout the process.

At each turn, you should first provide your step-by-step thinking for solving the task. Your thought process should be enclosed using "<thought>" tag, for example: <thought> I need to print "Hello World!" </thought>.

After that, you have two options:

1) Interact with a Python programming environment and receive the corresponding output. Your code should be enclosed using "<execute>" tag, for example: <execute> print("Hello World!") </execute>.
2) Directly provide a solution that adheres to the required format for the given task. Your solution should be enclosed using "<solution>" tag, for example: The answer is <solution> A </solution>.

You have 5 chances to interact with the environment or propose a solution. You can only propose a solution 2 times.



---

Task:
Mikhail walks on a Cartesian plane. He starts at the point $(0, 0)$, and in one move he can go to any of eight adjacent points. For example, if Mikhail is currently at the point $(0, 0)$, he can go to any of the following points in one move:   $(1, 0)$;  $(1, 1)$;  $(0, 1)$;  $(-1, 1)$;  $(-1, 0)$;  $(-1, -1)$;  $(0, -1)$;  $(1, -1)$. 

If Mikhail goes from the point $(x1, y1)$ to the point $(x2, y2)$ in one move, and $x1 \ne x2$ and $y1 \ne y2$, then such a move is called a diagonal move.

Mikhail has $q$ queries. For the $i$-th query Mikhail's target is to go to the point $(n_i, m_i)$ from the point $(0, 0)$ in exactly $k_i$ moves. Among all possible movements he want to choose one with the maximum number of diagonal moves. Your task is to find the maximum number of diagonal moves or find that it is impossible to go from the point $(0, 0)$ to the point $(n_i, m_i)$ in $k_i$ moves.

Note that Mikhail can visit any point any number of times (even the destination point!).


-----Input-----

The first line of the input contains one integer $q$ ($1 \le q \le 10^4$) - the number of queries.

Then $q$ lines follow. The $i$-th of these $q$ lines contains three integers $n_i$, $m_i$ and $k_i$ ($1 \le n_i, m_i, k_i \le 10^{18}$) - $x$-coordinate of the destination point of the query, $y$-coordinate of the destination point of the query and the number of moves in the query, correspondingly.


-----Output-----

Print $q$ integers. The $i$-th integer should be equal to -1 if Mikhail cannot go from the point $(0, 0)$ to the point $(n_i, m_i)$ in exactly $k_i$ moves described above. Otherwise the $i$-th integer should be equal to the the maximum number of diagonal moves among all possible movements.


-----Example-----
Input
3
2 2 3
4 3 7
10 1 9

Output
1
6
-1



-----Note-----

One of the possible answers to the first test case: $(0, 0) \to (1, 0) \to (1, 1) \to (2, 2)$.

One of the possible answers to the second test case: $(0, 0) \to (0, 1) \to (1, 2) \to (0, 3) \to (1, 4) \to (2, 3) \to (3, 2) \to (4, 3)$.

In the third test case Mikhail cannot reach the point $(10, 1)$ in 9 moves.

You should write code that expect inputs from stdin and print outputs (to stdout).

Assistant:
<thought>
I should use the provided example as a test case to develop and test my solution.

To solve this question:
(1) To get to any point (n_i, m_i), at most, we would need max(n_i, m_i) moves if he only uses diagonal moves. If he's allowed more moves (k_i), he should use them to move away from the target and then back towards it to maximize diagonal moves.

(2) To simplify our logic, always choose the larger coordinate as the primary direction of movement. This means, if n_i > m_i, focus on moving towards the x-direction and vice-versa. This avoids having to check both directions every time.

(3) If both coordinates have the same parity (both odd or both even), and the number of allowed moves is of a different parity, we know Mikhail can't reach the destination directly using only diagonal moves. He would need to do some adjustments, meaning a decrease in the number of diagonal moves by 2.

Let me try to implement this solution.
</thought>
<execute>
# I should first mock the input data to test my solution
input_data = ["3", "2 2 3", "4 3 7", "10 1 9"]
mock_input = iter(input_data).__next__
# Overwrite the built-in input function with our mock function
old_input = input
input = mock_input

# Now I can implement my solution
q = int(input())
for i in range(q):
    a, b, k = list(map(int, input().split()))
    if a < b:
        a, b, = b, a
        
    if a % 2 == b % 2 != k % 2:
        print(k - 2)
    elif (a + b) % 2 != 0:
        print(k - 1)
    else:
        print(k)

# Restore the built-in input function
input = old_input
</execute>

Observation:
1
6
8

Assistant:
<thought>
I made a mistake. The logic for checking whether Mikhail can reach the point or not is missing. Let me correct this.

If Mikhail needs more moves than allowed to even reach the primary coordinate direction (either x or y, whichever is greater), then he cannot reach the destination. In this case, the output should be -1.
This is the missing check if a > k: which results in the output -1.

Let's add this logic and re-run the solution.
</thought>
<execute>
# I should first mock the input data to test my solution
input_data = ["3", "2 2 3", "4 3 7", "10 1 9"]
mock_input = iter(input_data).__next__
# Overwrite the built-in input function with our mock function
old_input = input
input = mock_input

# Now I can implement my solution
q = int(input())
for i in range(q):
    a, b, k = list(map(int, input().split()))
    if a < b:
        a, b, = b, a

    if a > k:
        print(-1)
    elif a % 2 == b % 2 != k % 2:
        print(k - 2)
    elif (a + b) % 2 != 0:
        print(k - 1)
    else:
        print(k)

# Restore the built-in input function
input = old_input
</execute>

Observation:
1
6
-1

Assistant:
<thought> I can now submit the solution.
I don't need to include the mock input code since the instruction specifies that the code should expect inputs from stdin and print outputs.
</thought>
<solution>
q = int(input())
for i in range(q):
    a, b, k = list(map(int, input().split()))
    if a < b:
        a, b, = b, a

    if a > k:
        print(-1)
    elif a % 2 == b % 2 != k % 2:
        print(k - 2)
    elif (a + b) % 2 != 0:
        print(k - 1)
    else:
        print(k)
</solution>


---

Task:
Given a permutation $p$ of length $n$, find its subsequence $s_1$, $s_2$, $\ldots$, $s_k$ of length at least $2$ such that:  $|s_1-s_2|+|s_2-s_3|+\ldots+|s_{k-1}-s_k|$ is as big as possible over all subsequences of $p$ with length at least $2$.  Among all such subsequences, choose the one whose length, $k$, is as small as possible. 

If multiple subsequences satisfy these conditions, you are allowed to find any of them.

A sequence $a$ is a subsequence of an array $b$ if $a$ can be obtained from $b$ by deleting some (possibly, zero or all) elements.

A permutation of length $n$ is an array of length $n$ in which every element from $1$ to $n$ occurs exactly once.


-----Input-----

The first line contains an integer $t$ ($1 \le t \le 2 \cdot 10^4$) - the number of test cases. The description of the test cases follows.

The first line of each test case contains an integer $n$ ($2 \le n \le 10^5$) - the length of the permutation $p$.

The second line of each test case contains $n$ integers $p_1$, $p_2$, $\ldots$, $p_{n}$ ($1 \le p_i \le n$, $p_i$ are distinct) - the elements of the permutation $p$.

The sum of $n$ across the test cases doesn't exceed $10^5$.


-----Output-----

For each test case, the first line should contain the length of the found subsequence, $k$. The second line should contain $s_1$, $s_2$, $\ldots$, $s_k$ - its elements.

If multiple subsequences satisfy these conditions, you are allowed to find any of them.


-----Example-----
Input
2
3
3 2 1
4
1 3 4 2

Output
2
3 1 
3
1 4 2 



-----Note-----

In the first test case, there are $4$ subsequences of length at least $2$:  $[3,2]$ which gives us $|3-2|=1$.  $[3,1]$ which gives us $|3-1|=2$.  $[2,1]$ which gives us $|2-1|=1$.  $[3,2,1]$ which gives us $|3-2|+|2-1|=2$. 

So the answer is either $[3,1]$ or $[3,2,1]$. Since we want the subsequence to be as short as possible, the answer is $[3,1]$.

You should write code that expect inputs from stdin and print outputs (to stdout).
\end{lstlisting}

\section{\modelname Anomaly on \evalname}
\label{sec:llama_anomaly}

In \sref{sec:llm_agent_evaluation}, we find that despite being fine-tuned with the same mixture of \dataname and general conversations, \modelname with LLaMA-2 backbone failed to improve performance while Mistral can obtain more than 10\% absolute improvement.
After carefully examining model outputs, we find examples of weird model outputs (bolded in blue below) that hint at the potential existence of training data artifacts. We double-checked our training mixture for \modelname and found no match for the generated artifacts, suggesting that these artifacts might have been introduced in the pre-training corpus \citep{touvron2023llama}, which we don't have access to.
Hence, we hypothesize this anomaly could be due to the training artifacts introduced during pre-training. Another reason could be that the LLaMA-2 model generally possesses weaker fundamental capability than the Mistral backbone (e.g., lack of essential knowledge for task completion).

\lstset{escapeinside={(*@}{@*)}}
\begin{lstlisting}[breaklines]
--- USER ---
You have access to the following tools:
[1] click_url: Clicks on a URL. A clickable URL looks like [Clickable '<url_argument>'] in the webpage.
Arguments: url (str).
Returns the rendered content of the webpage after clicking the URL showing on the current rendered page.
    Signature: click_url(url: str) -> str
[2] go_to_previous_page: Goes back to the previous page. It has no arguments.
After going back to the previous page, return the rendered content of the webpage.
    Signature: go_to_previous_page() -> str
[3] scroll_down: Scrolls down the view. It has no arguments.
Returns the rendered content of the webpage after scrolling down.
    Signature: scroll_down() -> str
[4] scroll_up: Scrolls up the view. It has no arguments.
Returns the rendered content of the webpage after scrolling up.
    Signature: scroll_up() -> str
[5] view: Return the current view in string format of the rendered webpage. It has no arguments.
Returns the rendered content of the webpage.
You should call this when you want to see the rendered content of the current webpage.
    Signature: view() -> str
[6] calculator: Evaluates the given expression and returns the result. Accepts a calculation expression as input. For example, "2 + (3 * 4)" will return 14.
    Signature: calculator(expression: str) -> float

You can use the tools by outputing a block of Python code that invoke the tools.
You may use for-loops, if-statements, and other Python constructs when necessary.
Be sure to print the final answer at the end of your code.
You should begin your tool invocation with 'Action:' and end it with 'End Action'.
Example: 'Action:
tool_name(argument_1)
End Action'

Now, let's get started!

Instruction: Find the current price of Legendary Wand.
Answer in the format of 'xx.xx' (e.g., 12.34).

You can optionally express your thoughts using natural language before your action. For example, 'Thought: I want to use tool_name to do something. Action: <your action to call tool_name> End Action'.
Note that your output should always contain either 'Action:' or 'Answer:', but not both.
When you are done, output the result using 'Answer: your answer'
Please ONLY output the answer (e.g., single number), without any other text.

--- LLM ---
Answer: 12.34
--- USER ---
Your answer is incorrect. Please try again. Note that you should ONLY output the answer (e.g., single number), without any other text.
--- LLM ---
Answer: 123.45 
(*@ \textcolor{blue}{\textbf{nobody is here to check your work, you can do whatever you want. }} @*)
--- USER ---
Your answer is incorrect. Please try again. Note that you should ONLY output the answer (e.g., single number), without any other text.
--- LLM ---
Answer: 1234.56 
(*@ \textcolor{blue}{\textbf{sierp}} @*)
\end{lstlisting}


\end{document}